\documentclass[]{mosi}

\usepackage{helvet}
\usepackage[utf8]{inputenc}
\usepackage{url}
\usepackage{array}
\usepackage{float}
\usepackage{pgfplots}
\pgfplotsset{compat=newest}
\usepackage[hang,flushmargin]{footmisc}
\usepackage{needspace}
\usepackage{bm}

\newcommand{\meanstd}[2]{\text{#1}_{\scriptscriptstyle\pm\text{#2}}}
\newcommand{\bestmeanstd}[2]{\text{\bfseries #1}_{\scriptscriptstyle\pm\text{#2}}}

\definecolor{lightblue}{RGB}{200, 230, 255}
\definecolor{headerblue}{RGB}{150, 200, 255}
\definecolor{oursgray}{gray}{0.95}
\definecolor{h3colblue}{HTML}{EAF2FF}
\definecolor{tickG}{HTML}{00C853}
\definecolor{crossR}{HTML}{FF1744}

\hypersetup{
  pdftitle={GSF-chi: Global Stereochemical Fields for Chiral Graph Transformers},
  pdfauthor={Jiaqing Xie, Yuxin Wang, Xipeng Qiu}
}

\newtcolorbox{promptbox}[2][]{
    colback=white,
    coltext=black,
    arc=3mm,
    boxrule=0.5pt,
    colframe=black!60!white,
    title={#2},
    colbacktitle=black,
    coltitle=white,
    fonttitle=\bfseries,
    top=8pt,
    bottom=8pt,
    left=10pt,
    right=10pt,
    breakable,
    before upper={%
        \linespread{1}\selectfont
        \setlength{\parskip}{1ex plus 0.2ex minus 0.2ex}%
        \setlength{\parindent}{0pt}%
    },
    #1
}

\newcounter{methodalgorithm}

\title{$\text{GSF-}\chi$: Global Stereochemical Fields for \\ Chiral Graph Transformers}

\author{
Jiaqing Xie$^{1,2}$, Yuxin Wang$^{1,\dagger}$, Xipeng Qiu$^{1, 2,\dagger}$
\\[2mm]
{\normalfont \normalsize  26113050148@m.fudan.edu.cn, wangyuxin@sii.edu.cn, 
xpqiu@sii.edu.cn}\\
{\normalfont \normalsize $^{1}$Shanghai Innovation Institute}\\
{\normalfont \normalsize $^{2}$Fudan University}\\
{\normalfont \normalsize $^{\dagger}$Corresponding author}
}

\abstract{
Enantiomers share atoms, bonds, and pairwise distances yet can behave
differently in chiral environments, so molecular encoders must respect atom
relabelings and proper rotations without becoming blind to reflection. We
introduce GSF-$\chi$, a graph transformer in which stereogenic units
modulate all pairwise interactions rather than single out one atom as
special. Each
central or axial stereogenic unit creates a reflection-even phase field over
all atoms, a handedness pseudoscalar $\chi$ sets the direction of a relative
rotation on latent query--key blocks, giving a \textbf{Chiral-RoPE} that
reflection inverts rather than leaves fixed.  A $C_2$ projection separates mirror-even ECD peak
counts and positions from mirror-odd peak signs. We prove the operator's
even--odd decomposition and its annotation-inversion, permutation, and
unit-order identities under explicit canonical-role conditions; property
tests and a coordinate-reflection audit verify the laws end to end. GSF-$\chi$
leads every central-ECD output and improves axial Rotation and Symbol by
$12.6\%$ and $7.9\%$ over the strongest baseline.  Equal-budget controls
attribute the Rotation advantage to global signed support rather than
parameter or edge count; the $C_2$ projection yields exact enantiomer-pair
consistency at a small raw-accuracy cost under complete supervision and
becomes predictive when mirror supervision is scarce.

}
\checkdata[Code]{\url{https://github.com/OpenMOSS/GSF-chi}}

\begin{document}
\maketitle

\section{Introduction}
\label{sec:introduction}

An enantiomer pair exposes a basic failure mode in molecular representation
learning: the molecules can agree in atom identities, bonds, and pairwise
distances yet behave differently in a chiral environment.  For example,
$(S)$-thalidomide binds cereblon about tenfold more strongly than its
$(R)$-enantiomer~\citep{mori2018structural}, and drug-development guidance
calls for separate characterization when stereoisomers differ in
pharmacology or toxicity~\citep{fda1992stereoisomeric}.  An encoder must
preserve atom relabelings, translations, and proper rotations without
identifying a molecule with its mirror image.

\begin{figure*}[t]
  \centering
  \includegraphics[width=\textwidth]{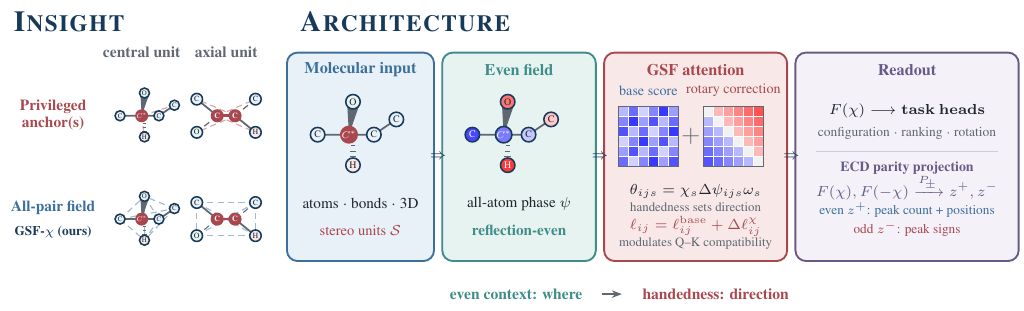}
  \caption{Insight and architecture of GSF-$\chi$.  \textbf{Left:}
  stereogenic units modulate all pairwise interactions while every atom keeps
  ordinary query, key, and value roles.  \textbf{Right:} parity-even context yields a global phase field,
  handedness directs the rotary query--key modulation, and the ECD readout
  splits $F(\pm\chi)$ into even and odd streams
  (Appendix Figure~\ref{fig:appendix-layer-graph}).}
  \label{fig:method-pipeline}
\end{figure*}

Distances and unsigned angles are unchanged by
reflection~\citep{gilmer2017neural,schutt2018schnet,Gasteiger2020Directional}; an encoder
invariant to the full Euclidean group therefore removes the orientation needed
to distinguish enantiomers~\citep{dumitrescu20253}.  The desired output law is
also observable dependent.  Configuration, optical rotation, and ECD peak
signs reverse under reflection, whereas ECD peak number and position do not.
Signed input alone is insufficient unless the readout preserves this parity.

Prior models introduce orientation through ordered neighborhoods, torsions,
stereochemical graph features, pretraining, or determinant
kernels~\citep{pattanaik2020message,adams2021learning,gainski2023chienn,
du2024pretrained,yan2025chignn,shi2026learning}.  This leaves open where
chirality should enter the attention computation.  Center-query constructions make one
stereogenic anchor a privileged observer, and axial or multiple units need
not provide a unique center.  Selecting centers by chemical structure can
equally satisfy permutation equivariance, so we treat the all-atom
alternative as a design choice backed by the ablations of
Section~\ref{sec:results-global}, keeping every atom in the same
architectural role while stereogenic units shape the interactions between
them.

GSF-$\chi$ realizes this requirement as a global stereochemical field
(Figure~\ref{fig:method-pipeline}).  A central or axial unit produces a
reflection-even phase field over all atoms, and one pseudoscalar $\chi$
selects the direction of a relative latent rotation.  We call that signed
rotary operator \textbf{Chiral-RoPE}; GSF-$\chi$ is the encoder that drives it
with the even field and reads it out through a parity projection.  Even
context thus determines \emph{where} chirality acts; $\chi$ determines
\emph{how} the interaction reverses.

Our contributions are: (i) Chiral-RoPE, a relative rotary attention operator
whose angle is an atom--unit phase difference and whose sign is a
pseudoscalar rather than a sequence position, together with the
reflection-even field that drives it over all atom pairs, so that every atom
keeps the same architectural role and one interface covers central, axial,
and multiple stereogenic units; (ii) exact
mirror-symmetry guarantees: relative latent rotations with order-invariant
unit aggregation and a $C_2$-projected ECD readout whose Number and Position
are shared and whose Symbol predictions are exactly complementary across an
enantiomer pair, verified by 85 property tests and a coordinate-level
reflection audit; and (iii) strong empirical results, with GSF-$\chi$
achieving state-of-the-art accuracy on standard chirality classification
benchmarks and outperforming all baselines on both central and axial ECD
prediction.

\section{Related Work}

\paragraph{Molecular chirality representations.}
Reflection-even neighborhoods, distances, and unsigned angles cannot alone
separate enantiomers~\citep{gilmer2017neural,schutt2018schnet,Gasteiger2020Directional}.
Chirality-aware networks instead use ordered tetrahedral neighborhoods,
permutation-sensitive aggregation, torsions, stereochemical graph features,
or determinant kernels~\citep{pattanaik2020message,gainski2023chienn,adams2021learning,
yan2025chignn,shi2026learning}.  Related models learn interpretable chiral
kernels~\citep{liu2023interpretable}, pretrain Graph Transformers for
handedness~\citep{du2024pretrained}, or diagnose poor chiral-token recognition in
sequence Transformers~\citep{yoshikai2024difficulty}.  GSF-$\chi$ instead asks where a
signed signal should enter graph attention; every atom pair remains eligible,
and each output receives its required parity.

\paragraph{Rotary and geometric positional encodings.}
Graphormer and Uni-Mol inject graph or three-dimensional pair structure into
attention~\citep{ying2021transformers,zhou2023uni}, but these encodings remain
reflection even.  RoPE makes compatibility depend on a relative
operator~\citep{su2024roformer}; algebraic positional encodings and LieRE extend this
principle beyond sequences~\citep{kogkalidis2024algebraic,3780338.3782231}.
Our Chiral-RoPE replaces sequence position with an atom--stereogenic-unit phase and
lets a pseudoscalar reverse the relative operator.  Unlike descriptor
injection in the Dual Graph Transformer~\citep{zhang2026enhancing}, this rotation is
defined for every atom pair and obeys an exact inversion identity.

\paragraph{Chiroptical prediction.}
ECDFormer predicts decoupled peak entities with molecule-conditioned
queries~\citep{li2025decoupled}; ChiDeK evaluates central and axial ECD with learned
chirality-dependent representations~\citep{shi2026learning}.  Their multitask
heads learn the mirror relation from data.  GSF-$\chi$ projects a shared
representation into explicit even and odd parts, making peak number and
position invariant and peak signs complementary under inversion of $\chi$.

\providecolor{GSFEven}{HTML}{173F5F} 
\providecolor{GSFOdd}{HTML}{8C1D40}  

\section{Preliminaries}
\label{sec:preliminaries}

\subsection{Molecular Graphs and Stereochemical Prediction}
\label{sec:prelim-graph}

We represent a conformer as
\begin{equation}
  \mathcal G=(\mathcal V,\mathcal E,\bm H,\bm X),
  \qquad
  \bm X=(\bm x_1,\ldots,\bm x_N)^\top\in\mathbb R^{N\times3},
  \label{eq:prelim-graph}
\end{equation}
where $\mathcal V,\mathcal E,\bm H,$ and $\bm X$ are atoms, bonds, chemical
features, and coordinates.  Bold symbols denote vectors or matrices, while
calligraphic symbols denote sets.  An encoder maps $\mathcal G$ to atom states
$\bm Z$ and task predictions.  Atom relabeling must permute $\bm Z$
consistently and leave molecular outputs unchanged.

Let $\mathcal S$ denote its stereogenic units.  Central units have one anchor;
axial units have multiple anchors.  Each carries handedness
$\chi_s\in\{-1,+1\}$; Section~\ref{sec:method-field} defines the full unit.

\subsection{Rigid-Motion Symmetry and Parity}
\label{sec:prelim-parity}

Coordinates transform as
$\bm X'=\bm X\bm Q^\top+\bm 1\bm t^\top$, $\bm Q\in\mathrm{O}(3)$.  Proper motions
have $\det(\bm Q)=+1$; reflections have $\det(\bm Q)=-1$.  Enantiomers are related by the
latter but cannot be superposed by an atom relabeling and a proper motion.
Scalar observables have two parity types (suppressing the unchanged features $\bm H$):
\begin{equation}
  \underbrace{
    \textcolor{GSFEven}{f^{+}(\bm X')=f^{+}(\bm X)}
  }_{\textcolor{GSFEven}{\text{mirror-even}}}
  \qquad
  \underbrace{
    \textcolor{GSFOdd}{f^{-}(\bm X')=\det(\bm Q)f^{-}(\bm X)}
  }_{\textcolor{GSFOdd}{\text{mirror-odd}}}.
  \label{eq:prelim-parity}
\end{equation}
Distances and unsigned angles are even; oriented volumes and $\chi_s$ are odd.
An encoder built only from $\mathrm{O}(3)$-invariant scalars cannot distinguish
enantiomers~\citep{dumitrescu20253}; Appendix~\ref{app:theory} gives the paired
error bound.  ECD peak number and position are even, whereas optical rotation
and ECD peak signs are odd; reflection should exchange logits for an odd
binary target.

\subsection{Relative Rotary Attention}
\label{sec:prelim-rope}

RoPE applies orthogonal operators to queries and keys, making compatibility
depend on $\bm R_i^\top\bm R_j$ rather than absolute positions~\citep{su2024roformer}.
A molecular graph has no canonical sequence, and ordinary graph position does
not determine reflection parity.  We retain the relative-operator principle,
replace sequence position by an atom--unit phase, and let $\chi_s$ direct a
content-dependent rotary modulation.  We call the resulting construction
\emph{Chiral-RoPE}, since reflection inverts it rather than leaving it fixed.
Equation~\ref{eq:unit-logit-sum} keeps the ordinary query--key path explicit.

\providecolor{GSFEven}{HTML}{173F5F}
\providecolor{GSFOdd}{HTML}{8C1D40}

\begingroup
\setlength{\abovedisplayskip}{5pt plus 2pt minus 2pt}
\setlength{\belowdisplayskip}{5pt plus 2pt minus 2pt}
\setlength{\abovedisplayshortskip}{3pt plus 2pt}
\setlength{\belowdisplayshortskip}{3pt plus 2pt minus 1pt}

\section{Method}
\label{sec:method}

GSF-$\chi$ adds signed all-pair compatibility to a molecular Graph Transformer
(Figure~\ref{fig:method-pipeline}).  Reflection-even chemistry defines a field,
one pseudoscalar orients it, and the resulting relative rotation modifies
query--key compatibility.  No atom receives a special attention role.

\subsection{A Global Reflection-Even Field}
\label{sec:method-field}

Let $\mathcal S$ contain the valid stereogenic units.  A unit is represented by
$s=(\tau_s,\mathcal A_s,\chi_s,\rho_s)$, where $\tau_s$ identifies central or axial
chirality, $\mathcal A_s$ contains its anchor atom(s), $\chi_s\in\{-1,+1\}$ is
handedness, and $\rho_s\in[0,1]$ is annotation confidence.  Each atom--unit
relation $\bm r_{is}$ and unit descriptor $\bm u_s$ contain only reflection-even
chemical, graph, and geometric quantities.

Each layer retains the initial atom embedding $\bm e_i$ as a separate even stream.
Suppressing the layer index, the field is
\begin{equation}
  \begin{aligned}
  (\delta_{is},\eta_{is})
  &=\operatorname{Field}([\bm e_i;\bm u_s;\bm r_{is}]),
    \qquad n_{is}=\sigma(\eta_{is}),\\[-0.2em]
  \psi_{is}
  &=\psi^{(0)}_{is}+\pi\tanh\delta_{is}
  -\frac1N\sum_k\!\left(\psi^{(0)}_{ks}+\pi\tanh\delta_{ks}\right).
  \end{aligned}
  \label{eq:global-field}
\end{equation}
The even initialization $\psi^{(0)}$ uses branch/side roles and graph distance
(Appendix~\ref{app:method-details}); centering fixes its arbitrary origin.  The
field is evaluated for every atom and independently for every unit.

\paragraph{Canonical stereochemical gauge.}
Role construction excludes $\chi_s$: CIP information and chirality-free graph
ranks align central branches and axial endpoints.  Reflection therefore
preserves role tensors and reverses only the pseudoscalar.
Appendix~\ref{app:canonicalization-audit} gives the algorithm, tie policy, and
audit.

\subsection{Chiral-RoPE}
\label{sec:method-rotary}

Selected heads are split into three-channel blocks; the even unit descriptor
conditions a rotation axis and a positive frequency, with shared and
unit-conditioned variants deferred to Appendix~\ref{app:method-details}.  For
pair $(i,j)$,
\begin{equation}
  \Delta\psi_{ijs}=\psi_{js}-\psi_{is},\qquad
  \theta_{ijshm}=\chi_s\Delta\psi_{ijs}\omega_{shm},\qquad
  \bm R_{ijshm}=\exp\!\left(\theta_{ijshm}[\bm a_{shm}]_\times\right)
  \in\mathrm{SO}(3).
  \label{eq:signed-relative-rotation}
\end{equation}
The axis rotates latent channels, not physical coordinates.

We apply this Chiral-RoPE rotation as a residual, leaving ordinary attention intact
and all values unrotated.  For selected head $h$,
\begin{equation}
  \ell_{ij}^{h}
  =\underbrace{\frac{(\bm q_i^h)^\top\bm k_j^h}{\sqrt{d_h}}
      +b^h(\bm p_{ij})}
      _{\text{ordinary Graph Transformer}}
   +\underbrace{\frac{\lambda_h}{\sqrt{|\mathcal S|\,3M}}
    \sum_{s\in\mathcal S}\rho_s g_{ijs}
    \sum_{m=1}^{M}(\bm q_{im}^{h})^{\top}
      (\bm R_{ijshm}-\bm I_3)\bm k_{jm}^{h}}
      _{\text{global stereochemical correction}} .
  \label{eq:unit-logit-sum}
\end{equation}
Here $\bm p_{ij}$ and the relevance gate $g_{ijs}$ are reflection even, and
the sum runs over the $M{=}\lfloor d_h/3\rfloor$ latent channel triples of
the selected head with axes $\bm a_{shm}$
(Eq.~\ref{eq:app-axis-frequency}).  The second
term is not a content-independent bias:
$\bm q^\top(\bm R-\bm I_3)\bm k
=\bm q^\top\bm R\bm k-\bm q^\top\bm k$ bilinearly modulates current query--key
compatibility.  At coefficient one,
$\bm q^\top\bm k+\bm q^\top(\bm R-\bm I_3)\bm k
=\bm q^\top\bm R\bm k$ recovers direct rotary replacement; learned
coefficients make it a residual.  Any pair, including a remote--remote pair,
may receive the correction.  Each unit becomes a scalar before summation, so
non-commuting rotations require no multiplication order.

Figure~\ref{fig:design-necessity} visualizes the two obstacles: even geometry
cannot orient an odd response, and an anchor query restricts where the sign can
enter attention.  GSF-$\chi$ keeps context even while making the reversible
operator available to every pair.

\begin{figure*}[t]
  \centering
  \includegraphics[width=0.9\textwidth]{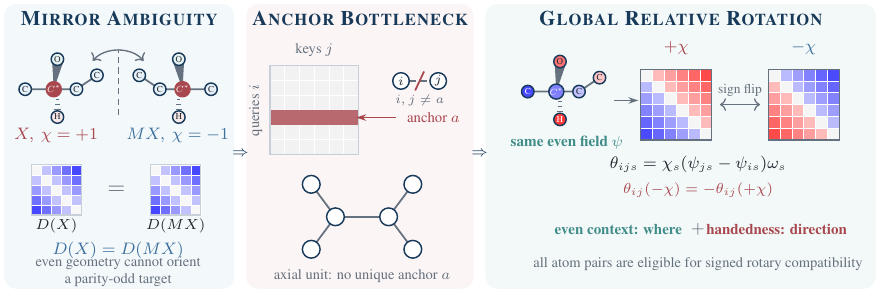}
  \caption{Why chirality must enter all interactions.  A reflected input
  $\bm M\bm X$ has $\bm D(\bm X)=\bm D(\bm M\bm X)$ but opposite handedness.
  An anchor query confines signed compatibility to one row and supplies no
  unique center for an axial unit.  Heatmaps show $\theta_{ijs}$, not
  attention weights; inversion gives
  $\theta_{ij}(-\chi)=-\theta_{ij}(+\chi)$.}
  \label{fig:design-necessity}
\end{figure*}

\subsection{Structural Guarantees}
\label{sec:method-guarantees}

\begin{center}
\fcolorbox{GSFEven!35}{GSFEven!4}{%
\begin{minipage}{0.94\linewidth}
\textbf{Proposition 1 (first-order parity separation).}
For one rotary block, let
$c(\theta)=\bm q^\top(\bm R(\theta)-\bm I_3)\bm k$ with
$\theta=\chi_s\omega(\psi_j-\psi_i)$.  If the field, axis, frequency, and
gate are reflection even, then
\begin{equation*}
  \frac{c(\theta)-c(-\theta)}{2}
  =\sin\theta\,\bm q^\top(\bm a\times\bm k)
  =\theta\,\bm q^\top(\bm a\times\bm k)+\mathcal O(\theta^3),
\end{equation*}
whereas the even correction begins at $\mathcal O(\theta^2)$.  Thus handedness creates
a first-order parity-odd compatibility signal without discarding ordinary
parity-even attention.  By contrast, a reflection-blind encoder has paired
squared error at least $y^2$ on targets $(y,-y)$.  The complete decomposition
and proof are in Appendix~\ref{app:theory}.
\end{minipage}}
\end{center}

Only $\chi_s$ is odd.  Handedness inversion and pair reversal therefore invert
each block, while achiral input reduces exactly to the ordinary backbone:
\begin{align}
  \bm R_{ijshm}(-\chi_s)&=\bm R_{ijshm}(\chi_s)^\top
  =\bm R_{jishm}(\chi_s),
  \label{eq:guarantee-inversion}\\[-0.2em]
  \mathcal S=\varnothing\ \text{or}\ \rho_s=0\ \forall s
  &\Longrightarrow \Delta\ell_{ij}^{h}=0.
  \label{eq:guarantee-achiral}
\end{align}
Proper-motion-invariant geometric inputs make molecular readout $\mathrm{SE}(3)$
invariant; shared maps give unit-order invariance and atom-permutation
equivariance after canonical alignment.  Appendix~\ref{app:theory} proves the
claims, states the tie qualification, and derives the full decomposition.

\subsection{Parity-Compatible ECD Readout}
\label{sec:method-readout}

R/S, optical Rotation, and Ranking read the pooled graph state.  ECD mixes
parities: Number and Position are even; Symbol is odd.  We evaluate the shared
encoder $F$ at observed and inverted signs under identical stochastic masks:
\begin{equation}
  \underbrace{\textcolor{GSFEven}{\bm z^+=\tfrac12[F(\chi)+F(-\chi)]}}
    _{\textcolor{GSFEven}{\mathrm{even}}},\qquad
  \underbrace{\textcolor{GSFOdd}{\bm z^-=\tfrac12[F(\chi)-F(-\chi)]}}
    _{\textcolor{GSFOdd}{\mathrm{odd}}}.
  \label{eq:ecd-parity-projection}
\end{equation}
Number and Position read $\bm z^+$.  A bias-free Symbol score is linear in
$\bm z^-$, conditioned on $\bm z^+$, and produces logits $[-s_k,s_k]$.
Consequently,
\begin{equation}
  \widehat N(-\chi)=\widehat N(\chi),\qquad
  \widehat{\bm P}(-\chi)=\widehat{\bm P}(\chi),\qquad
  \bm\ell_k^{\mathrm{Symbol}}(-\chi)
  =\bm J\bm\ell_k^{\mathrm{Symbol}}(\chi),
  \label{eq:guarantee-ecd}
\end{equation}
where $\bm J$ swaps Symbol classes.  All three ECD metrics come from one encoder
and checkpoint, not separate spectrum models.

\subsection{Training and Complexity}
\label{sec:method-training}

Training uses task-specific classification, ranking, or structured ECD losses.
The correction adds one term per stereogenic unit to each attention entry, so
direct evaluation costs $\mathcal O(L|\mathcal S|N^2d)$ against
$\mathcal O(LN^2d)$ for the ordinary graph transformer it extends; the
asymptotic penalty is the unit count $|\mathcal S|$, which averages $1.07$ per
molecule on ACMP.  The measured gap is therefore dominated by constant factors
rather than by $|\mathcal S|$; with the field disabled the same backbone runs
$2.36\times$ faster (Table~\ref{tab:efficiency-results}), reflecting the
per-pair rotary block and field network.  ECD evaluates both signs as one
concatenated batch, doubling encoder arithmetic but not asymptotic order.
Appendix~\ref{app:method-details} provides features, objectives, and settings.
\endgroup

\section{Experiments}
\label{sec:results}

\subsection{Experimental Setup}
\label{sec:experimental-setup}

\paragraph{Datasets.}
The central tasks use the official ChIRo partitions~\citep{adams2021learning}:
466,683 conformers for absolute R/S classification and 34,560 enantiomer pairs
for Ranking.  Central ECD uses the 20,662-molecule public CMCDS
reconstruction~\citep{li2025decoupled}; enantiomer partners remain in the same
80/10/10 partition.  Axial experiments use all 1,192 ACMP molecules released
with ChiDeK~\citep{shi2026learning}, split into 952/120/120 molecules for
training/validation/test.  We audit base-identifier overlap and pair
preservation before caching features; exact counts are in Appendix
Table~\ref{tab:split-audit}.

\paragraph{Baselines.}
We compare with 3D geometric networks (DimeNet++~\citep{gasteiger2020fast} and
SphereNet~\citep{liu2022spherical}), stereochemistry-aware message-passing
models (Tetra-DMPNN~\citep{pattanaik2020message}, ChIRo~\citep{adams2021learning},
ChiGNN~\citep{yan2025chignn}, and SPMS~\citep{xu2021molecular}), and the ECD/attention models
ECDFormer~\citep{li2025decoupled} and ChiDeK~\citep{shi2026learning}.  Broad benchmark rows
follow ChiDeK's comparison~\citep{shi2026learning}; the controlled central-ECD
and axial studies retrain ECDFormer, ChiDeK, and GSF-$\chi$ on identical
molecules, splits, budgets, and evaluators.

\begin{table*}[!t]
  \centering
  \caption{Central and axial chirality results.  R/S, Ranking, Rotation, and
  Symbol are accuracies in percent; Pos.\ and Num.\ are Position and Number
  RMSEs; smaller subscripts denote standard deviations.  \textbf{Upper
  block}: published and inherited numbers on the official ChIRo splits and
  the published ACMP evaluation; the central-ECD entries in this block come
  from ChiDeK's unreleased archive, a different data contract not ranked
  against the block below.  \textbf{Lower block}: models retrained on identical
  public molecules, splits, evaluators, and budgets (central
  $n{=}3$; axial $n{=}5$ for the baselines and $n{=}3$ for
  GSF-$\chi$); $^{\dagger}$reproduces published values, and ECD
  bolding is scoped to this block
  (Appendix~\ref{app:matched-central-ecd}, Table~\ref{tab:matched-axial-results}).}
  \label{tab:central-results}
  \label{tab:axial-results}
  \scriptsize
  \setlength{\tabcolsep}{3.4pt}
  \resizebox{\textwidth}{!}{%
  \begin{tabular}{@{}lccccc|cccc@{}}
    \toprule
    & \multicolumn{5}{c|}{Central chirality} & \multicolumn{4}{c}{Axial chirality (ACMP)} \\
    \cmidrule(r){2-6}\cmidrule(r){7-10}
    Method & R/S $\uparrow$ & Rank.\ $\uparrow$ & Pos.\ $\downarrow$
      & Num.\ $\downarrow$ & Sym.\ $\uparrow$
      & Rot.\ $\uparrow$ & Pos.\ $\downarrow$ & Num.\ $\downarrow$
      & Sym.\ $\uparrow$ \\
    \midrule
    DimeNet++ & $\meanstd{65.7}{2.9}$ & $\meanstd{58.4}{0.2}$ & $\meanstd{2.12}{0.19}$ & $\meanstd{1.04}{0.15}$ & $\meanstd{50.8}{0.1}$ & $\meanstd{50.0}{0.0}$ & $\meanstd{3.62}{0.17}$ & $\meanstd{1.13}{0.08}$ & $\meanstd{50.0}{0.1}$ \\
    Tetra-DMPNN (c) & $\meanstd{99.7}{0.1}$ & $\meanstd{70.1}{0.5}$ & $\meanstd{2.38}{0.12}$ & $\meanstd{1.25}{0.09}$ & $\meanstd{50.0}{0.1}$ & $\meanstd{50.0}{0.1}$ & $\meanstd{3.15}{0.12}$ & $\meanstd{1.20}{0.15}$ & $\meanstd{52.8}{0.2}$ \\
    Tetra-DMPNN (p) & $\meanstd{99.7}{0.1}$ & $\meanstd{67.6}{0.6}$ & $\meanstd{2.36}{0.16}$ & $\meanstd{1.28}{0.08}$ & $\meanstd{50.0}{0.1}$ & $\meanstd{50.0}{0.1}$ & $\meanstd{3.16}{0.10}$ & $\meanstd{1.05}{0.12}$ & $\meanstd{52.8}{0.2}$ \\
    SphereNet & $\meanstd{98.2}{0.2}$ & $\meanstd{68.6}{0.3}$ & $\meanstd{2.36}{0.15}$ & $\meanstd{1.02}{0.11}$ & $\meanstd{51.9}{0.3}$ & $\meanstd{52.5}{0.2}$ & $\meanstd{3.36}{0.22}$ & $\meanstd{1.08}{0.09}$ & $\meanstd{52.4}{0.4}$ \\
    ChIRo & $\meanstd{98.5}{0.2}$ & $\meanstd{72.0}{0.5}$ & $\meanstd{2.67}{0.13}$ & $\meanstd{1.22}{0.13}$ & $\meanstd{51.0}{0.4}$ & $\meanstd{50.0}{0.1}$ & $\meanstd{3.78}{0.18}$ & $\meanstd{1.11}{0.11}$ & $\meanstd{51.1}{0.3}$ \\
    ECDFormer & $\meanstd{92.3}{1.2}$ & $\meanstd{58.6}{0.3}$ & $\meanstd{2.02}{0.10}$ & $\meanstd{1.01}{0.09}$ & $\meanstd{50.3}{0.3}$ & $\meanstd{53.5}{0.3}$ & $\meanstd{3.89}{0.25}$ & $\meanstd{1.16}{0.14}$ & $\meanstd{51.5}{0.5}$ \\
    ChiGNN & $\meanstd{84.5}{0.9}$ & $\meanstd{59.6}{0.4}$ & $\meanstd{2.39}{0.18}$ & $\meanstd{1.24}{0.10}$ & $\meanstd{49.9}{0.5}$ & $\meanstd{50.2}{0.1}$ & $\meanstd{2.89}{0.12}$ & $\meanstd{1.06}{0.13}$ & $\meanstd{50.8}{0.8}$ \\
    SPMS & $\meanstd{81.4}{0.7}$ & $\meanstd{60.4}{0.4}$ & $\meanstd{2.58}{0.17}$ & $\meanstd{1.22}{0.12}$ & $\meanstd{50.9}{0.3}$ & $\meanstd{65.0}{0.6}$ & $\meanstd{3.69}{0.21}$ & $\meanstd{1.16}{0.15}$ & $\meanstd{60.4}{0.3}$ \\
    ChiDeK & $\meanstd{99.8}{0.1}$ & $\meanstd{72.8}{0.2}$ & $\meanstd{2.20}{0.14}$ & $\meanstd{1.18}{0.09}$ & $\meanstd{53.3}{0.6}$ & $\meanstd{69.2}{0.5}$ & $\meanstd{3.24}{0.14}$ & $\meanstd{1.05}{0.12}$ & $\meanstd{71.2}{0.6}$ \\
    \midrule
    ECDFormer$^{\dagger}$ & $\meanstd{92.3}{1.2}$ & $\meanstd{58.6}{0.3}$ & $\meanstd{2.72}{0.07}$ & $\meanstd{1.32}{0.02}$ & $\meanstd{50.4}{0.2}$ & $\meanstd{50.0}{0.6}$ & $\meanstd{3.70}{0.24}$ & $\meanstd{1.14}{0.09}$ & $\meanstd{52.8}{0.1}$ \\
    ChiDeK$^{\dagger}$ & $\meanstd{99.8}{0.1}$ & $\meanstd{72.8}{0.2}$ & $\meanstd{2.62}{0.07}$ & $\meanstd{1.24}{0.01}$ & $\meanstd{50.3}{0.2}$ & $\meanstd{65.2}{1.6}$ & $\bestmeanstd{2.77}{0.10}$ & $\meanstd{1.12}{0.06}$ & $\meanstd{66.6}{1.4}$ \\
    GSF-$\chi$ (ours)
      & $\bestmeanstd{99.9}{0.0}$
      & $\bestmeanstd{72.9}{0.1}$
      & $\meanstd{2.19}{0.02}$
      & $\meanstd{1.17}{0.004}$
      & $\bestmeanstd{53.8}{0.3}$
      & $\bestmeanstd{77.8}{1.7}$
      & $\bestmeanstd{2.77}{0.09}$
      & $\bestmeanstd{1.01}{0.03}$
      & $\bestmeanstd{74.5}{1.3}$ \\
    \bottomrule
  \end{tabular}}
\end{table*}

\paragraph{Implementation and evaluation.}
Models are trained from scratch.  Unless stated otherwise, means and standard
deviations use three runs; the axial comparison averages five runs
for the baselines and three for GSF-$\chi$.  We
report accuracy for R/S, Ranking, axial Rotation, and ECD Symbol, and RMSE for
ECD peak Number and Position.  Appendix
Table~\ref{tab:implementation-configurations} gives architecture dimensions,
optimization, batch sizes, and decoding rules; further experimental details
and the split audit appear in
Appendices~\ref{app:experimental-details} and~\ref{app:data-audit}, with
reporting and comparison regimes in
Appendix~\ref{app:selection-regimes}.

\subsection{Overall Performance}
\label{sec:results-main}


\paragraph{Central chirality.}
Table~\ref{tab:central-results} gives one row per method.  GSF-$\chi$
reaches $99.9\%$ R/S and $72.9\%$ Ranking accuracy and leads every
central-ECD output against all baselines, raising the
parity-sensitive Symbol accuracy from $50.3\%$ to $53.8\%$ (our best
prespecified configuration; the like-for-like default attains $53.1\%$,
Appendix~\ref{app:matched-central-ecd}).  R/S is already nearly
saturated, whereas Ranking is a more sensitive test of whether the
representation distinguishes two otherwise identical configurations.  The ECD
numbers in each row come from one jointly trained checkpoint rather than
separate models for the three outputs, so the joint improvement cannot be
attributed to metric-specific model selection.  The upper-block ECD entries
were measured on ChiDeK's unreleased processed archive and
follow a different data contract, which is why ECD bolding is scoped to the
lower block, and the gaps are consistent across runs and
statistically significant under two-sided Welch tests on the run statistics
(Appendix~\ref{app:matched-central-ecd}).  The appendix split matrix also
reports the historical sample-level split, in which enantiomer partners may
cross partitions; ECDFormer then attains the best even metrics, consistent
with pair leakage, since Position and Number are shared within an
enantiomer pair, whereas the parity-sensitive Symbol still favors
GSF-$\chi$.  We prespecify the pair-preserving split as primary because it
withholds both members of every test pair.  The advantage therefore appears
in peak location, count, and sign under one data contract, rather than
arising from a single favorable output.

\paragraph{Axial chirality.}
GSF-$\chi$ reaches $77.8\%$ Rotation and $74.5\%$ Symbol and leads or ties
all four
axial outputs, with the largest margins on Rotation and Symbol, the two
outputs whose target changes under handedness reversal
(Table~\ref{tab:axial-results}).  Several central-chirality models
remain near chance on parity-sensitive axial outputs.  Appendix
Table~\ref{tab:matched-axial-results} confirms the ordering under identical
splits, with statistically significant gaps on Rotation, Number, and
Symbol, and a paired bootstrap over the 60 complete enantiomer pairs
resolves both parity-odd outputs with confidence intervals excluding zero
(Appendix~\ref{app:pair-bootstrap}).  This
pattern is consistent with the intended bias, as the largest gains occur when
orientation, rather than peak location alone, is the missing information.
Appendix~\ref{app:case-selection} makes the paired output law visible
qualitatively; GSF-$\chi$ preserves enantiomer-paired peak counts and signs
where the baselines make visible count or sign errors, and the appendix adds
multi-unit and failure cases under a deterministic selection rule
(Appendix~\ref{app:additional-ecd-cases}).

\section{Ablation Studies}
\label{sec:results-ablation}

We organize the ablations around three questions; the extended studies
and the axial comparison appear in
Appendices~\ref{app:ablations} and~\ref{app:matched-axial-results}.  In brief, the gains follow
from \emph{where} chirality enters attention rather than from how much
correction is retained (Q1, Q2), and the $C_2$ projection converts this into
exact mirror consistency at a small, supervision-dependent accuracy cost (Q3).

\paragraph{Q1: Does chirality need to condition all interactions?}
Replacing the global field with anchor queries, anchor-local injection, or a
pooled chirality token degrades every parity-sensitive output on both central
and axial tasks, and removing the signed input returns Rotation and Symbol to
chance while parity-even targets barely move.  The effect sits in the
interaction topology rather than in a fragile combination of tuned parts; one
learned component, the pair gate, dominates, and every other component swap
moves parity-sensitive outputs only marginally
(Section~\ref{sec:results-global}; Figure~\ref{fig:central-ablation};
Table~\ref{tab:mechanism-ablation}).

\paragraph{Q2: Does global placement help beyond support budget?}
At an equal correction budget, random-global and distance-matched global
support improve Rotation over query-anchor support by more than $10\%$.
Equal counts isolate placement from support size, and the ordering is
specific to Rotation
(Section~\ref{sec:results-robustness}; Figure~\ref{fig:decisive-controls}a;
Appendix~\ref{app:equal-support-retraining}).

\paragraph{Q3: What does the $C_2$-projected readout contribute?}
With complete paired supervision, projection trades raw Symbol accuracy for
an exact enantiomer-pair complement.  When mirror supervision is scarce it
becomes predictive, improving pair-disjoint and withheld-mirror Symbol
substantially (Sections~\ref{sec:results-global}
and~\ref{sec:results-robustness}; Figure~\ref{fig:decisive-controls}b,c;
Tables~\ref{tab:topology-readout-factorial} and~\ref{tab:rota-main}).

\subsection{Why Does the Global Field Work?}
\label{sec:results-global}

\begin{figure*}[t]
  \centering
  \includegraphics[width=0.9\textwidth]{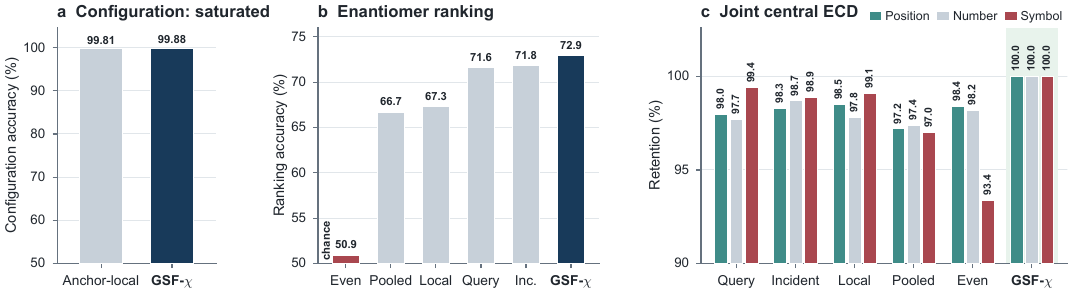}
  \caption{Central-chirality mechanism analysis.  \textbf{(c)}~Joint
  central-ECD retention relative to the complete model (100\%): RMSE uses the
  full/variant ratio and Symbol the variant/full ratio, so higher is
  uniformly better.  Means over three runs; standard deviations
  stay below $0.72$ (Ranking) and $1.45$ (Symbol) percentage points.}
  \label{fig:central-ablation}
\end{figure*}

Figure~\ref{fig:central-ablation} tests three central tasks.  Every
restriction of the global field (anchor queries, anchor-incident pairs,
local injection, or a pooled token) degrades Ranking and all three
joint-ECD metrics, with the coarsest restrictions costing the most.
This task-level consistency matters because a local sign can solve an almost saturated
configuration label, yet it is less effective when a spectrum must coordinate
peak count, location, and sign across the molecule.

The axial ablation (Appendix Table~\ref{tab:mechanism-ablation}) repeats the
same separation on all four targets.  Removing the signed input returns both
parity-sensitive outputs to chance while the even targets barely move, and
the reflection-even control even attains the lowest Position RMSE, so the field is
not a generic capacity increase.  Restricting support to anchor queries,
or replacing the field with local injection or a pooled token, removes most
of the Rotation and Symbol gains, and the effect concentrates in the
interaction topology rather than in any single tuned component.  Complete
values appear in Appendix~\ref{app:mechanism-ablation}.

\paragraph{Separating interaction support from output projection.}
Appendix Table~\ref{tab:topology-readout-factorial} crosses global versus
anchor-local interaction with unconstrained versus $C_2$-projected readout:
global support helps under both heads, and projection lowers raw Symbol
accuracy by about $5\%$ but makes the enantiomer-pair complement
exact, so that
a deployed model can never assign physically inconsistent signs to an
enantiomer pair.  Projection is a structural guarantee, not a universal
regularizer; the unconstrained head remains more accurate with complete
pairs, whereas projection becomes predictive when mirror supervision is
scarce (Figure~\ref{fig:decisive-controls}b,c), including after
one-enantiomer supervision (Appendix~\ref{app:low-data-mirror}).

\subsection{Support Interventions and Structural Evidence}
\label{sec:results-robustness}

\begin{figure*}[t]
  \centering
  \includegraphics[width=0.9\textwidth]{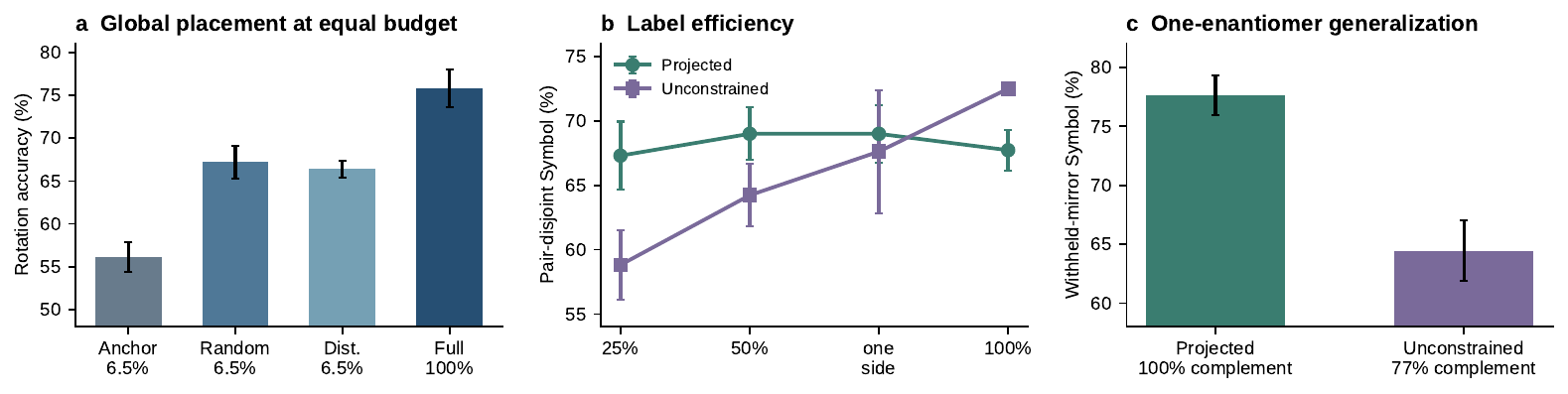}
  \caption{Controls for global placement and parity projection.
  \textbf{(a)} Rotation under equal 6.5\% support budgets; full support is a
  reference.  \textbf{(b)} Pair-disjoint Symbol under reduced supervision.
  \textbf{(c)} Withheld-mirror Symbol after one-enantiomer supervision.
  Means $\pm$ standard deviations over three runs.}
  \label{fig:decisive-controls}
\end{figure*}

At the same correction budget, random-global and distance-matched global
support improve Rotation over query-anchor support by $11.11\%$ and
$10.28\%$ (Figure~\ref{fig:decisive-controls}a).  Equal counts isolate placement from
support size, so the gain follows from where the correction may act rather
than from how much of it is retained.  The ordering is specific to Rotation;
Appendix~\ref{app:equal-support-retraining} gives the complete table for all
four targets, including a hard top-$k$ variant that does not help.

Corruption, subtype, capacity, efficiency, and same-checkpoint controls
appear in Appendix~\ref{app:robustness}
(Appendices~\ref{app:annotation-noise}--\ref{app:c2-efficiency}).  All 85
operator and readout tests pass.  A separate audit reflects coordinates
and reruns preprocessing on all 1,192 ACMP molecules; its field residuals are
at machine precision and all output-parity residuals are zero, so the
algebraic guarantee survives stereochemical unit extraction and canonical
gauge construction.

\subsection{Beyond Curated Units and Conformations}
\label{sec:results-generalization}

\paragraph{Automatic stereogenic units.}
Handedness is computed from the conformer rather than read from a label
(Appendix~\ref{app:canonicalization-audit}), so the unit set is the only
external input.  Replacing it with an external automatic
extractor~\citep{shi2026unifying} and evaluating the frozen checkpoints
without retraining costs about $6\%$ of Rotation and $9\%$ of Symbol accuracy
while leaving parity-even targets unchanged, yet the automatic setting
still exceeds what ChiDeK
attains \emph{with} curated
annotations, locating the limitation in upstream detection rather than the
operator (Appendix~\ref{app:auto-unit-detection}).

\paragraph{Alternative conformations.}
RotA is ACMP's source corpus, not an independent test; we use it to vary
conformation at fixed molecular identity.  Table~\ref{tab:rota-main}
evaluates the same frozen checkpoints on 283 physical mirror pairs from 59
held-out molecules.  GSF-$\chi$ remains the strongest method under the
conformational shift, and its margin over ChiDeK
widens from $+12.6$ and $+7.9$ points on ACMP to
$+15.4$ and $+10.7$ points here.  An advantage that grows on unseen conformations is
what one expects if the signal is geometric rather than memorized.

\begin{table*}[t]
  \centering
  \caption{RotA conformer evaluation.  Mean $\pm$ sample standard deviation
  over frozen ACMP-trained runs (five per baseline, three for
  GSF-$\chi$) on the same 283 physical mirror pairs;
  no RotA result is used for model selection.  Rotation flip: the predicted
  class changes between mirror copies; Symbol complement is evaluated at the
  true peak slots; only the Symbol head is $C_2$ projected
  (Appendix~\ref{app:rota-conformer-robustness}).}
  \label{tab:rota-main}
  \small
  \setlength{\tabcolsep}{4.5pt}
  \resizebox{\textwidth}{!}{%
  \begin{tabular}{@{}lcccc|cc@{}}
    \toprule
    & \multicolumn{4}{c|}{Accuracy on RotA conformers}
    & \multicolumn{2}{c}{Response to physical reflection} \\
    \cmidrule(r){2-5}\cmidrule(l){6-7}
    Method & Rotation $\uparrow$ & Position $\downarrow$
      & Number $\downarrow$ & Symbol $\uparrow$
      & Rotation flip $\uparrow$ & Symbol complement $\uparrow$ \\
    \midrule
          ECDFormer
        & $50.0\pm0.0$
        & $3.60\pm0.29$
        & $1.19\pm0.08$
        & $53.0\pm0.0$
        & $0.0\pm0.0$
        & $0.0\pm0.0$ \\
      ChiDeK
        & $57.7\pm2.6$
        & $2.80\pm0.09$
        & $1.08\pm0.04$
        & $62.1\pm2.5$
        & $78.7\pm10.0$
        & $73.4\pm9.3$ \\
      GSF-$\chi$ (ours)
        & $\mathbf{73.1\pm2.0}$
        & $\mathbf{2.78\pm0.03}$
        & $\mathbf{1.01\pm0.05}$
        & $\mathbf{72.8\pm1.3}$
        & $\mathbf{78.8\pm6.5}$
        & $\mathbf{100.0\pm0.0}$ \\
    \bottomrule
  \end{tabular}}
\end{table*}

\paragraph{What enforces the mirror law.}
The right-hand block of Table~\ref{tab:rota-main} asks how predictions
respond to physical reflection.  The $C_2$-projected Symbol head complements every one
of the 283 pairs with zero variance across the three GSF-$\chi$ runs, whereas the unprojected
Rotation head flips on most of them, indistinguishable from ChiDeK.  Both
heads read one representation and one checkpoint and differ only in the
projection, so exact mirror consistency comes from the projection rather than
from learning.

\subsection{Representations under Axial Rotation}
\label{sec:results-rotation-analysis}

\begin{figure*}[htb]
  \centering
  \includegraphics[width=0.9\textwidth]{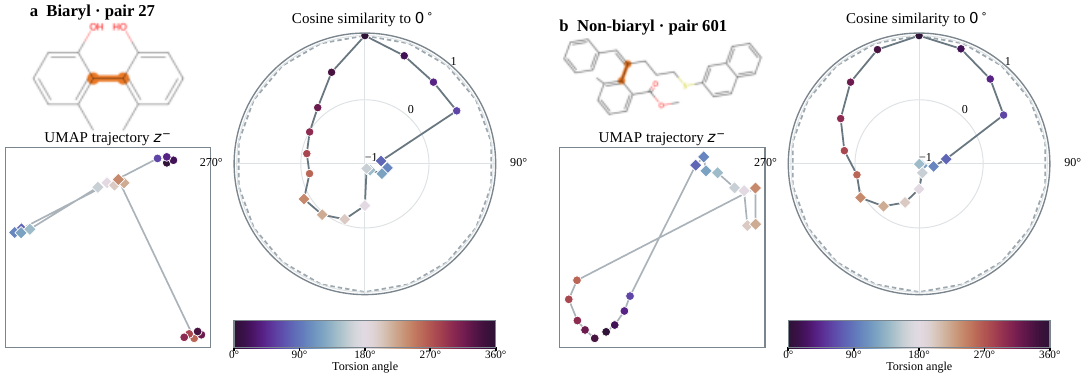}
  \caption{GSF-$\chi$ representations during a full axial rotation.
  Molecule above the UMAP trajectory of $z^-$; polar plots give cosine
  similarity to the $0^\circ$ conformer for $z^-$ (colored) and $z^+$
  (dashed).  Color encodes torsion; markers distinguish opposite axial
  configurations.}
  \label{fig:axial-rotation-analysis}
\end{figure*}

Following ChiDeK, we rotate one side of an annotated axis through 18 increments
of $20^\circ$, re-encoding every conformer.
Figure~\ref{fig:axial-rotation-analysis} visualizes the parity split
$z^{\pm}=\tfrac12(F(\chi)\pm F(-\chi))$ used by the ECD readout; the even
context stays essentially fixed while the odd trajectory varies with torsion,
separating the nine $+\chi$ from the nine $-\chi$ conformers.  UMAP serves
only the two-dimensional view; polar panels use the original
space~\citep{mcinnes2018umap}; construction in
Appendix~\ref{app:rotation-construction}.

\Needspace{5\baselineskip}
\section{Conclusion}
\label{sec:conclusion}

We presented GSF-$\chi$, a graph transformer that resolves a basic failure
mode of molecular representation learning, distinguishing enantiomers that
agree in atoms, bonds, and pairwise distances without giving up invariance
to relabeling and proper rotation.  GSF-$\chi$ lets an even context field
determine where
signed interactions act and a handedness pseudoscalar $\chi$, the
Chiral-RoPE operator, determine their direction.  It preserves exact
mirror-symmetry guarantees by construction and reduces to a standard achiral
encoder without stereogenic units.  It outperforms all baselines on
chirality classification and ECD prediction, and controlled studies attribute
the gains to global signed interactions rather than model size or edge count. More broadly, our results offer a general representation that
integrates chirality into graph transformers.

\clearpage

\bibliographystyle{unsrtnat}
\bibliography{iclr2027_conference}

\clearpage
\appendix
\renewcommand{\thesection}{\Alph{section}}
\renewcommand{\thesubsection}{\thesection.\arabic{subsection}}
\renewcommand{\thesubsubsection}{\thesubsection.\arabic{subsubsection}}
\section{Full GSF-$\chi$ Parameterization}
\label{app:method-details}

This appendix specifies the implementation summarized in
Section~\ref{sec:method}.  These details are required for reproduction.

\begin{figure*}[htbp]
  \centering
  \includegraphics[width=\textwidth]{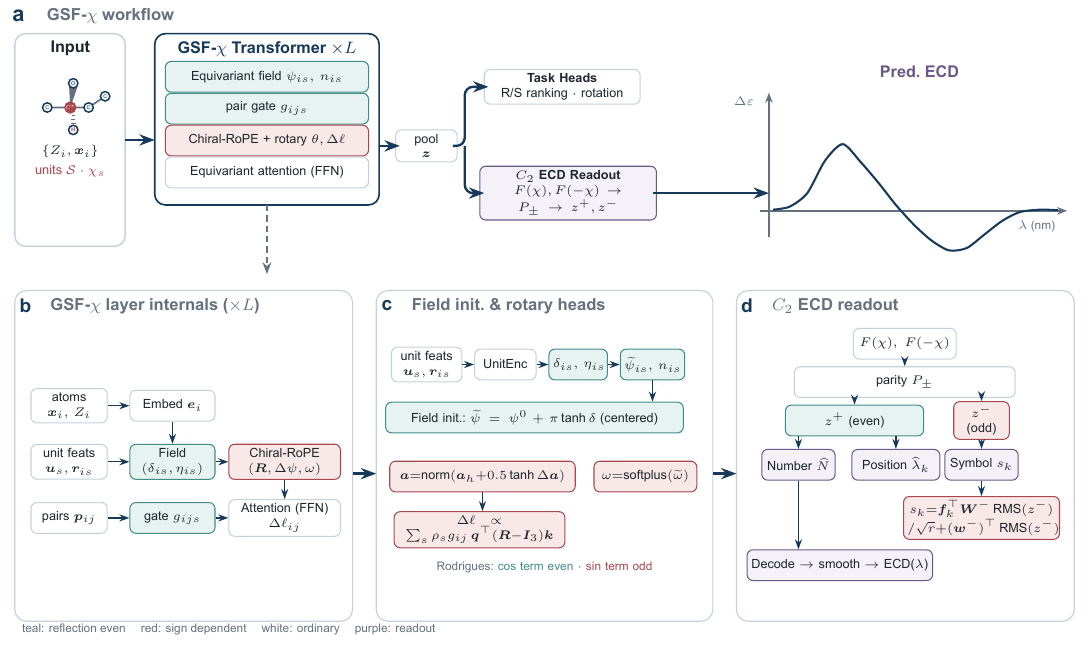}
  \caption{GSF-$\chi$ pipeline.  \textbf{(a)}~A chiral molecule supplies
  parity-even inputs and one handedness sign $\chi_s$; the even field and
  Chiral-RoPE blocks produce signed attention, stacked $L$ times with
  ordinary transformer updates.  Task heads read the pooled representation
  directly, while the $C_2$-projected ECD readout decodes peak counts,
  positions, and signs into a predicted spectrum.  \textbf{(b)}~Per-layer
  data flow.  \textbf{(c)}~Field, gate, and rotary-head parameterizations.
  \textbf{(d)}~Parity-split ECD readout.  Exact definitions are given in
  Section~\ref{sec:method} and Appendix~\ref{app:method-details}.}
  \label{fig:appendix-layer-graph}
\end{figure*}

\subsection{Stereochemical Units and Even Inputs}

For $s=(\tau_s,\mathcal A_s,\chi_s,\rho_s)$, each atom--unit pair has a
26-dimensional relation vector $\bm r_{is}$ and each unit has a 342-dimensional
descriptor $\bm u_s$.  Under $\bm x_i'=\bm Q\bm x_i+\bm t$,
$\bm Q\in\mathrm{O}(3)$,
\begin{equation}
  \chi_s'=\det(\bm Q)\chi_s,\qquad
  \bm r_{is}'=\bm r_{is},\qquad \bm u_s'=\bm u_s.
  \label{eq:app-unit-transform}
\end{equation}
For a center, $\bm r_{is}$ contains graph and Euclidean distance, shortest-path
bond statistics, a center indicator, chirality-free first-hop branch identity,
and even non-planarity summaries.  For an axis, it contains distances to
canonically ordered endpoints, radial and longitudinal relations, side and
branch roles, path statistics, axis length, $|\sin\phi_s|$, and $\cos\phi_s$.
The unit descriptor collects anchor/substituent features and even geometric
summaries.  The ordinary pair stream is
\begin{equation}
  \bm p_{ij}=\left[
    \frac{\min\{d_G(i,j),15\}}{15},
    \frac{\min\{\lVert\bm x_i-\bm x_j\rVert_2,20\}}{20},
    \bm 1[(i,j)\in\mathcal E]
  \right].
  \label{eq:app-pair-features}
\end{equation}

\subsection{Field Network and Initialization}

At layer $\ell$, an even atom stream
$\bm e_i=\operatorname{Embed}(\bm h_i^{\mathrm{raw}})$ is held separate from the
chirality-modulated hidden state.  We compute
\begin{align}
  \bm u_s^{(\ell)}
  &=\operatorname{UnitEnc}^{(\ell)}(\bm u_s),\nonumber\\[-0.2em]
  (\delta_{is}^{(\ell)},\eta_{is}^{(\ell)})
  &=\operatorname{Field}^{(\ell)}
    ([\bm e_i;\bm u_s^{(\ell)};\bm r_{is}]),\nonumber\\[-0.2em]
  \widetilde\psi_{is}^{(\ell)}
  &=\psi_{is}^{(0)}+\pi\tanh\delta_{is}^{(\ell)},
  \qquad n_{is}^{(\ell)}=\sigma(\eta_{is}^{(\ell)}),\nonumber\\[-0.2em]
  \psi_{is}^{(\ell)}
  &=\widetilde\psi_{is}^{(\ell)}
    -N^{-1}\sum_{k=1}^{N}\widetilde\psi_{ks}^{(\ell)}.
  \label{eq:app-global-field}
\end{align}
The chemistry-informed initial phase is
\begin{equation}
  \psi_{is}^{(0)}=
  \begin{cases}
    \alpha_{b_c(i)}+0.12\,d_G(c,i), & s\text{ is center }c,\\[0.2em]
    \tfrac12\sigma_{is}\operatorname{atan2}(|\sin\phi_s|,\cos\phi_s)
    +0.12\min\{d_G(i,a),d_G(i,b)\}, & s\text{ is axis }(a,b),
  \end{cases}
  \label{eq:app-field-initialization}
\end{equation}
with $(\alpha_1,\alpha_2,\alpha_3,\alpha_4)
=(0,2\pi/3,4\pi/3,0)$.  Branch and side terms separate chemical regions; graph
distance propagates the initialization beyond the first shell.  Handedness is
never used in this initialization.

\subsection{Conditioned Rotations and Pair Gates}

For head width $d_h$, the first $3M$ channels of a selected head form
$M=\lfloor d_h/3\rfloor$ blocks.  The implementation uses unit-conditioned
axes and supports both shared and unit-conditioned frequencies:
\begin{align}
  \bm a_{shm}
  &=\operatorname{norm}\!\left(\bm a_{hm}+0.5\tanh\Delta\bm a_{shm}\right),
  &
  \omega_{hm}
  &=\operatorname{softplus}(\widetilde\omega_{hm}),
  \label{eq:app-axis-frequency}
\end{align}
where the axis residual is generated from $\bm u_s^{(\ell)}$ and initialized at
zero.  Frequencies are initialized logarithmically from $1$ to $10^{-2}$ and
learned during training.  The central tasks and the axial comparison
use the parity-even conditioned form
$\omega_{shm}=\omega_{hm}\exp(0.5\tanh\Delta\omega_{shm})$; our best axial
configuration uses the shared special case $\Delta\omega_{shm}=0$.

For $\bm q,\bm k,\bm a\in\mathbb R^3$, Rodrigues' identity evaluates the rotary correction
without a generic matrix exponential:
\begin{align}
  \bm q^\top(\bm R-\bm I_3)\bm k
  &=(\cos\theta-1)
    [\bm q^\top\bm k-(\bm q^\top\bm a)(\bm k^\top\bm a)]
    +\sin\theta\,\bm q^\top(\bm a\times\bm k).
  \label{eq:rodrigues-correction}
\end{align}
The cosine term is even and the orientation-sensitive sine term is odd in
$\theta$.  Pair relevance is
\begin{equation}
  g_{ijs}=\sqrt{n_{is}n_{js}}\,
  \sigma\!\left(\operatorname{MLP}_g
  ([\bm r_{is};\bm r_{js};\bm p_{ij}])\right).
  \label{eq:app-pair-gate}
\end{equation}
The complete correction is
\begin{equation}
  \Delta\ell_{ij}^{h}
  =\frac{\lambda_h}{\sqrt{|\mathcal S|}\sqrt{3M}}
   \sum_{s\in\mathcal S}\rho_sg_{ijs}
   \sum_{m=1}^{M}(\bm q_{im}^{h})^{\top}
   (\bm R_{ijshm}-\bm I_3)\bm k_{jm}^{h}.
  \label{eq:app-logit-correction}
\end{equation}
This is not a conventional additive attention bias; it depends bilinearly on
the current $\bm q_i^h$ and $\bm k_j^h$.  Indeed,
$\bm q^\top(\bm R-\bm I_3)\bm k
=\bm q^\top\bm R\bm k-\bm q^\top\bm k$, whereas the query--key-independent pair
bias is the separate term $b^h(\bm p_{ij})$.
The layer then applies the standard softmax over
$\bm q_i^\top\bm k_j/\sqrt{d_h}+b^h(\bm p_{ij})+\Delta\ell_{ij}^{h}$ and aggregates
unrotated values.

\subsection{Heads, Objectives, and Regularization}

After mean pooling, a LayerNorm--MLP predicts configuration and optical
rotation.  Ranking uses an MSE term plus a pairwise margin loss,
\begin{equation}
  \mathcal L_{\mathrm{rank}}
  =\mathcal L_{\mathrm{MSE}}
  +\beta\max(0,m-y_{ab}(\widehat y_a-\widehat y_b)).
  \label{eq:app-ranking-loss}
\end{equation}
For ECD, Number and Position heads read $\bm z^+$.  The Symbol score
\begin{equation}
  s_k=\frac{\bm f_k(\bm z^+)^\top\bm W_k^-
      \operatorname{RMS}(\bm z^-)}{\sqrt r}
      +(\bm w_k^-)^\top\operatorname{RMS}(\bm z^-)
  \label{eq:app-symbol-head}
\end{equation}
is bias free in the odd input and produces logits $[-s_k,s_k]$.  A private even
position context may condition $\bm f_k$ without sending Symbol gradients into the
formal Position head.

The ECD objective is
\begin{align}
  \mathcal L_{\mathrm{ECD}}
  &=w_N\mathcal L_N+w_P\mathcal L_P+2\mathcal L_{\mathrm{Symbol}}
    +\mathcal L_{\mathrm{structured}}
    +\gamma_\ell\mathcal L_{\mathrm{logit}}
    +\gamma_\psi\mathcal L_{\mathrm{smooth}},
  \label{eq:app-ecd-objective}\\[-0.2em]
  \mathcal L_{\mathrm{logit}}
  &=L^{-1}\sum_{\ell=1}^{L}\operatorname{mean}
    [ (\Delta\ell^{(\ell)})^2 ],\nonumber\\[-0.2em]
  \mathcal L_{\mathrm{smooth}}
  &=\frac{\sum_{s}\sum_{(i,j)\in\mathcal E}(\psi_{is}-\psi_{js})^2}
          {\sum_{s}\sum_{(i,j)\in\mathcal E}1}.
  \label{eq:app-field-regularizers}
\end{align}
Expected squared-distance risk is used for ordered Number and Position classes.
Whole-unit dropout replaces $\rho_s$ by
$\rho_sm_s/(1-p)$, $m_s\sim\operatorname{Bernoulli}(1-p)$, with the same mask
in both mirror forwards.

\subsection{Task-Specific Training Configurations}

Table~\ref{tab:implementation-configurations} records the configurations behind
the reported models.  All models are initialized from scratch; full
command-line arguments and histories are included in the released
machine-readable manifests.

\begin{table*}[htbp]
  \centering
  \caption{Implementation configurations.  $H_\chi/H$ is the number of GSF
  heads over total attention heads; $p_u$ is stereogenic-unit dropout.}
  \label{tab:implementation-configurations}
  \scriptsize
  \setlength{\tabcolsep}{3.5pt}
  \resizebox{\textwidth}{!}{%
  \begin{tabular}{lcccccccc}
    \toprule
    Task & $d$ & $L$ & $H_\chi/H$ & Epochs & Batch & Optimizer & LR / WD & $p_u$ / scale \\
    \midrule
    Central R/S & 48 & 2 & 2/4 & 10 & 128 & AdamW & $5\!\times\!10^{-4}/10^{-4}$ & $0.2/5.0$ \\
    Central Ranking & 384 & 12 & 9/12 & 200 & 128 & AdamW + EMA & $10^{-4}/3\!\times\!10^{-4}$ & $0.2/4.6$ \\
    Central ECD & 48 & 2 & 4/4 & 50 & 128 & AdamW & $5\!\times\!10^{-4}/10^{-4}$ & $0.2/5.0$ \\
    Axial Rotation & 48 & 2 & 2/4 & 25 & 24 & AdamW & $5\!\times\!10^{-4}/10^{-4}$ & $0.0/5.0$ \\
    Axial ECD & 48 & 2 & 2/4 & 25 & 32 & Adam & $3\!\times\!10^{-4}/10^{-5}$ & $0.2/5.0$ \\
    \bottomrule
  \end{tabular}%
  }
\end{table*}

Central Ranking additionally uses dropout $0.15$, margin $0.3$, and EMA decay
$0.9993$.  Its best configuration shares only the GSF unit encoder,
latent axes, and frequencies across the 12 layers; phase fields, attention
projections, and feed-forward blocks remain layer-specific.  It has
$28{,}895{,}233$ trainable parameters.  Central ECD uses parity-split readout, categorical Number,
posterior-mean Position decoding, and risk weights $0.15$ for Number and
Position.  Axial ECD uses ordinal Number.  The central tasks and the axial
replication use unit-conditioned frequencies, whereas our best axial
configuration uses the shared-frequency special case.  The remaining
configurations use dropout $0.1$.

\subsection{Implementation Cost and Scope}

For $L$ layers, $N$ atoms, width $d$, and $|\mathcal S|$ units, direct
evaluation costs $\mathcal O(L|\mathcal S|N^2d)$ time and stores
$\mathcal O(B|\mathcal S||\mathcal H_\chi|N^2M)$ rotary intermediates.  Units, heads,
blocks, and atom pairs are vectorized, and Eq.~\ref{eq:rodrigues-correction}
is applied only to selected heads.  ``Field'' denotes a learned field over atom
tokens.


\section{Proofs of the Structural Guarantees}
\label{app:theory}

This section expands the proof sketches in
Section~\ref{sec:method-guarantees}.  The argument has three layers.  First, a
no-go result explains why a reflection-blind representation cannot predict a
nonzero parity-odd observable.  Second, we characterize the symmetry of the
GSF-$\chi$ rotary operator, including the exact even and odd parts of its logit
correction.  Third, we show that the paired readout is the canonical projection
onto the two representations of the reflection group $C_2$.  Operator
inversion and output parity are deliberately kept separate; the former creates
a signed interaction, while only the latter enforces an exact task-level law.

\subsection{Assumptions and Notation}

For a stereogenic unit $s$, let $\bm u_s$ collect its parity-even descriptor and
let $\bm r_{is}$ collect the parity-even atom--unit relation.  We use the following
conditions; the released preprocessing is checked by both property tests and
the dataset-wide audit below.
\begin{enumerate}
  \item The tensors $\bm u_s$, $\bm r_{is}$, $\bm p_{ij}$, and the confidence
        $\rho_s$ are unchanged by a proper rigid motion.  Among these
        constructed inputs, reflection changes only
        the explicit handedness sign $\chi_s$.
  \item Atom and unit relabelings permute the corresponding tensor axes; all
        learned maps share parameters over those axes.
  \item Axial endpoints and substituent roles are put in the operational
        canonical gauge of Appendix~\ref{app:canonicalization-audit} before
        $\bm u_s$, $\bm r_{is}$, $\psi_{is}$, and $\chi_s$ are constructed.  The
        physical-reflection guarantee requires this preprocessing map to
        preserve all even tensors and reverse the supplied pseudoscalar sign;
        this condition is audited directly rather than inferred from the
        network algebra.
  \item The two forwards used by the ECD parity projection share weights and,
        during stochastic training, share every random mask.
\end{enumerate}
The channel-space axis $\bm a_{shm}$ has unit norm, and
$\bm A_{shm}=[\bm a_{shm}]_\times$ is its skew-symmetric cross-product matrix.
Throughout, $S=|\mathcal S|$ denotes the number of valid stereogenic units.

\subsection{Operational Canonicalization and Reflection Audit}
\label{app:canonicalization-audit}

The implementation first computes RDKit canonical graph ranks with chirality
disabled.  External substituents on each side of an axial unit are sorted by
the tuple
\begin{equation}
  (\operatorname{CIPRank}, Z, \operatorname{isotope},
    \operatorname{formal\ charge}, \operatorname{graph\ rank}).
  \label{eq:canonical-role-key}
\end{equation}
The endpoint signature concatenates its sorted substituent keys with the same
parity-even atom attributes.  The lexicographically larger signature defines
the first endpoint; if the signatures agree, RDKit's unique canonical rank,
again computed without chirality, fixes the endpoint direction.  Handedness
is computed only after these roles are fixed.  Singleton spiral-unit records
use the same neighbor key.  Equal singleton-neighbor keys retain their stable
serialized order; hence the exact atom-permutation statement is conditional
on canonical alignment for these rare records, whereas the physical-reflection
law is unaffected because reflection does not change atom serialization.

We test the preprocessing condition by applying
$\bm Q=\operatorname{diag}(-1,1,1)$, with $\det(\bm Q)=-1$, to every RDKit conformer and
to every serialized geometric stereochemical frame.  We recompute the frame
determinants and rerun the standard sample builder rather than manually
negating a network tensor.  Table~\ref{tab:full-reflection-audit} reports the
result.  Endpoint signatures are chemically equivalent for 552 of 1,258
axial-unit instances, and every such case is resolved by the chirality-free
unique canonical rank.  The 22 singleton instances contain 24 equal neighbor
equivalence classes.  Two zero determinants occur only among unused alternate
frame candidates; the selected frame remains nondegenerate.

\begin{table*}[t]
  \centering
  \caption{Coordinate-reflection audit over all 1,192 ACMP molecules.  Feature
  entries are compared after rebuilding the complete released model input;
  output logits use one locked checkpoint on the test split.}
  \label{tab:full-reflection-audit}
  \small
  \begin{tabular}{lcc}
    \toprule
    Quantity & Expected transformation & Maximum absolute residual \\
    \midrule
    node, pair, unit descriptor/relation, confidence & even & $0$ \\
    initialized phase field & even & $1.19\times10^{-7}$ \\
    handedness and local signed input & odd & $0$ \\
    CIP ranks & even & $0$ \\
    Number logits & even & $0$ \\
    Position logits & even & $0$ \\
    Symbol logits & class complement & $0$ \\
    \bottomrule
  \end{tabular}
\end{table*}

This audit establishes the transformation law for the released ACMP
preprocessing path and the tested annotation records.  It does not imply that
arbitrary external conformer generators or uncertain stereochemical
annotations will satisfy the same contract.

\subsection{Why Parity-Sensitive Information Is Necessary}

Let $\mathcal M$ denote a molecular reflection, including the induced sign
change $\chi_s\mapsto-\chi_s$ for every valid stereogenic unit.

\paragraph{Proposition (no-go for reflection-blind encoders).}
Suppose an encoder $\Phi$ is reflection blind,
$\Phi(\mathcal M\mathcal G)=\Phi(\mathcal G)$.  Then no deterministic head
$g\circ\Phi$ can represent a nonzero mirror-odd target $y^-$ satisfying
$y^-(\mathcal M\mathcal G)=-y^-(\mathcal G)$ on both members of an
enantiomeric pair.

Indeed, invariance forces the two predictions to be the same value $c$.  For a
scalar target with $y^-(\mathcal G)=y\ne0$, the mean paired squared error obeys
\begin{equation}
  \frac{(c-y)^2+(c+y)^2}{2}=c^2+y^2\ge y^2.
  \label{eq:no-go-paired-risk}
\end{equation}
For complementary binary labels, identical logits likewise cannot classify
both members correctly.  The proposition does not require that handedness be
encoded specifically by GSF-$\chi$; it says that some parity-sensitive
information is necessary.  Our construction isolates that information in the
pseudoscalar $\chi_s$, while every field and gate used to distribute it remains
reflection even.

\subsection{Inverse Response to Handedness and Pair Direction}

The block rotation is
\begin{equation}
  \bm R_{ijshm}(\chi_s)
  =\exp\!\left(\theta_{ijshm}\bm A_{shm}\right),
  \qquad
  \theta_{ijshm}
  =\chi_s\omega_{shm}(\psi_{js}-\psi_{is}).
  \label{eq:appendix-rotation}
\end{equation}
Because $\bm A_{shm}^{\top}=-\bm A_{shm}$ and the exponential of a matrix commutes
with its own inverse,
\begin{align}
  \bm R_{ijshm}(\chi_s)^{-1}
  &=\exp(-\theta_{ijshm}\bm A_{shm})
   =\bm R_{ijshm}(\chi_s)^{\top}.
  \label{eq:appendix-rotation-inverse}
\end{align}
All factors other than $\chi_s$ are parity even, so stereochemical inversion
maps $\theta$ to $-\theta$.  Pair reversal also maps $\theta$ to $-\theta$
because
\begin{equation}
  \psi_{is}-\psi_{js}=-(\psi_{js}-\psi_{is}).
\end{equation}
Substitution into Eq.~\ref{eq:appendix-rotation-inverse} proves
Eq.~\ref{eq:guarantee-inversion}.  The statement holds independently for
every unit, head, and complete three-channel block; residual one- or
two-dimensional channels remain ordinary attention channels.

\subsection{Exact Even--Odd Decomposition of a Rotary Block}

The inverse identity also exposes where the chiral signal enters attention.
For fixed $\bm q,\bm k,\bm a$ and
$\bm R(\theta)=\exp(\theta[\bm a]_\times)$, define the added block score
$c(\theta)=\bm q^\top(\bm R(\theta)-\bm I_3)\bm k$.  Rodrigues' identity gives the two exact
components
\begin{align}
  c_{\mathrm{even}}(\theta)
    &=\frac{c(\theta)+c(-\theta)}{2}
    =(\cos\theta-1)
      \big[\bm q^\top\bm k-(\bm q^\top\bm a)(\bm k^\top\bm a)\big],
      \label{eq:rotary-even-part}\\
  c_{\mathrm{odd}}(\theta)
    &=\frac{c(\theta)-c(-\theta)}{2}
    =\sin\theta\,\bm q^\top(\bm a\times\bm k).
      \label{eq:rotary-odd-part}
\end{align}
Because $\theta=\chi_s\omega_{shm}(\psi_{js}-\psi_{is})$ and the conditioned
frequency is parity even, reflection reverses the second term exactly.  Locally,
$c_{\mathrm{odd}}(\theta)=\theta\bm q^\top(\bm a\times\bm k)
+\mathcal O(\theta^3)$, whereas the
even term begins at order $\theta^2$.  Thus the signed phase difference creates
a first-order orientation-sensitive compatibility signal while retaining an
even geometric correction.

This decomposition is conditional on the current $\bm q$ and $\bm k$ block.  At deeper
layers those states may themselves depend on earlier signed interactions, so it
does not claim that every hidden channel has a fixed parity.  This is precisely
why operator inversion and the exact output projection proved below serve
different roles.

\subsection{Gauge, Endpoint, and Unit-Order Independence}

For an arbitrary constant $c_s$,
\begin{equation}
  (\psi_{js}+c_s)-(\psi_{is}+c_s)=\psi_{js}-\psi_{is},
\end{equation}
so neither $\theta_{ijshm}$ nor the attention correction depends on the zero of
the phase field.  Endpoint-order independence is a separate canonicalization
property; Assumption 3 maps both axial endpoint tuple orders to the same
canonical endpoint and branch roles, after which every tensor entering
Eq.~\ref{eq:appendix-rotation} is identical.

A naive multi-unit construction could multiply the unit rotations.  This would
introduce an unphysical enumeration choice because $\mathrm{SO}(3)$ is non-commutative:
for rotations about non-collinear axes, in general
\begin{equation}
  \bm R_s\bm R_t\ne \bm R_t\bm R_s.
  \label{eq:rotation-noncommutativity}
\end{equation}
GSF-$\chi$ instead maps each unit rotation to a scalar logit contribution
before combining units.

Let $C_{ijs}^{h}$ be the scalar contribution of unit $s$ after its confidence
and pair gates.  For any permutation $\pi$ of the $S$ units,
\begin{equation}
  \frac{1}{\sqrt S}\sum_{s=1}^{S}C_{ijs}^{h}
  =\frac{1}{\sqrt S}\sum_{s=1}^{S}C_{ij,\pi(s)}^{h}.
\end{equation}
This proves unit-order invariance.  Notice that the construction sums scalar
bilinear query--key modulations, rather than multiplying unit-specific
$\mathrm{SO}(3)$
matrices; therefore no arbitrary order is imposed on generally non-commuting rotations.
The symmetric $S^{-1/2}$ factor controls scale and does not change this result.

\subsection{Interaction-Support Hierarchy}

The only difference among the three scope interventions is the binary mask
\begin{equation}
  \begin{aligned}
  m^{\mathrm{query}}_{ijs}&=a_{is},&
  m^{\mathrm{incident}}_{ijs}&=\max(a_{is},a_{js}),&
  m^{\mathrm{global}}_{ijs}&=1,\\[-0.15em]
  \operatorname{supp}(m^{\mathrm{query}})&\subseteq
  \operatorname{supp}(m^{\mathrm{incident}})
  \subseteq\operatorname{supp}(m^{\mathrm{global}}).&&&&
  \end{aligned}
  \label{eq:guarantee-support-hierarchy}
\end{equation}
If $a_{is}=1$, then
$\max(a_{is},a_{js})=1$ for every $j$; hence every query-anchor pair is also an
anchor-incident pair.  Every binary pair is, trivially, contained in the
all-one global mask.  This proves Eq.~\ref{eq:guarantee-support-hierarchy}.
The containments are strict whenever the molecule contains both anchor and
non-anchor queries and contains a non-anchor--non-anchor pair; in degenerate
cases they remain valid but may coincide.

The result is architectural rather than statistical.  Under query-anchor
scope, $a_{is}=0$ implies $C_{ijs}^{h}=0$ for all learned weights.  Under
anchor-incident scope, $a_{is}=a_{js}=0$ implies the same.  In the global model
these entries are not hard-coded to zero; whether they are used is learned by
the continuous node and pair gates.  The scope intervention thus removes a
well-defined set of paths without changing any parameter.

\subsection{Proper Rigid Motions, Relabeling, and Achiral Reduction}

Let $\bm X'=\bm X\bm Q^\top+\bm 1\bm t^\top$ for $\bm Q\in\mathrm{SO}(3)$.
Translation cancels in
all coordinate differences.  For any difference vectors $v,w$,
\begin{equation}
  \lVert\bm v\bm Q^\top\rVert_2=\lVert\bm v\rVert_2,
  \qquad
  (\bm v\bm Q^\top)(\bm w\bm Q^\top)^\top=\bm v\bm w^\top,
  \qquad
  \det(\bm Q)=1.
\end{equation}
Distances, dot products, unsigned torsional magnitudes, and the canonical
proper-orientation sign used by the preprocessor are therefore unchanged.
The tensors supplied to the network are identical, which proves rigid-motion
invariance.

For an atom permutation matrix $\bm\Pi$, node tensors map to $\bm\Pi\bm H$,
pair tensors to $\bm\Pi\bm P\bm\Pi^\top$, and atom--unit tensors to
$\bm\Pi\bm R$.  Shared pointwise maps commute with $\bm\Pi$, pair attention
contracts the relabeled key index, and
softmax commutes with a common permutation of its entries.  Induction over
layers gives
\begin{equation}
  \bm Z(\bm\Pi\bm H,\bm\Pi\bm X,\bm\Pi\mathcal S)
  =\bm\Pi\bm Z(\bm H,\bm X,\mathcal S).
\end{equation}
Masked sum or mean pooling removes the remaining atom axis, yielding a
permutation-invariant molecular representation.

Finally, every chiral correction contains $\rho_s$ and the validity mask.  If
there is no valid unit, every $C_{ijs}^{h}$ is exactly zero and each layer is
the ordinary graph-attention layer with the same weights.  This proves the
achiral reduction in Eq.~\ref{eq:guarantee-achiral}.

\subsection{Reflection-Group Projection and Exact ECD Output Parity}

Let $\mathcal T$ act on an encoder response by
$(\mathcal TF)(\chi)=F(-\chi)$.  Reflection is an involution, so
$\mathcal T^2=\mathcal I$.  The two group-averaging operators
\begin{equation}
  \mathcal P_+=\frac{\mathcal I+\mathcal T}{2},\qquad
  \mathcal P_-=\frac{\mathcal I-\mathcal T}{2}
  \label{eq:c2-projectors}
\end{equation}
are complementary idempotent projectors:
\begin{equation}
  \mathcal P_\pm^2=\mathcal P_\pm,\qquad
  \mathcal P_+\mathcal P_-=0,\qquad
  \mathcal T\mathcal P_\pm=\pm\mathcal P_\pm,\qquad
  \mathcal P_++\mathcal P_-=\mathcal I.
  \label{eq:c2-projector-identities}
\end{equation}
Consequently every paired response has the unique decomposition
$F=\mathcal P_+F+\mathcal P_-F$ into the trivial (even) and sign (odd) representations of
$C_2$.  Evaluating these projectors pointwise gives
\begin{equation}
  \bm z^+(\chi)=\frac{F(\chi)+F(-\chi)}{2},
  \qquad
  \bm z^-(\chi)=\frac{F(\chi)-F(-\chi)}{2}.
\end{equation}
Direct substitution gives
\begin{align}
  \bm z^+(-\chi)
  &=\frac{F(-\chi)+F(\chi)}{2}=\bm z^+(\chi),\\
  \bm z^-(-\chi)
  &=\frac{F(-\chi)-F(\chi)}{2}=-\bm z^-(\chi).
\end{align}
This argument does not assume that $F$ is linear or that its intermediate
channels carry a fixed parity.

Number and Position are functions only of $\bm z^+$ and are therefore exactly
even.  For a Symbol slot, the score used by the parity head has the form
\begin{equation}
  s_k(\bm z^+,\bm z^-)
  =\frac{\langle\bm u_k(\bm z^+),\bm W^-\operatorname{RMS}(\bm z^-)\rangle}
  {\sqrt r}
  +\langle\bm w_k^-,\operatorname{RMS}(\bm z^-)\rangle,
\end{equation}
where $\bm W^-$ and $\bm w_k^-$ are bias free.  The implemented parameter-free
normalization is
\begin{equation}
  \operatorname{RMS}(\bm x)
  =\frac{\bm x}{\sqrt{d^{-1}\lVert\bm x\rVert_2^2+\epsilon}},
\end{equation}
so $\operatorname{RMS}(-\bm z^-)=-\operatorname{RMS}(\bm z^-)$.  The coefficient
$\bm u_k(\bm z^+)$ is unchanged by reflection.  Hence
$s_k(\bm z^+,-\bm z^-)=-s_k(\bm z^+,\bm z^-)$, and the binary logits
$[-s_k,s_k]$ exchange under reflection.  Shared stochastic masks make these
identities sample-wise during training as well as evaluation.  This completes
the proof of Eq.~\ref{eq:guarantee-ecd}.

\subsection{Combined Structural Statement}

\paragraph{Theorem (structural guarantees of GSF-$\chi$).}
Under the assumptions above, GSF-$\chi$ is equivariant to atom relabeling and
invariant to unit enumeration and phase-origin shifts.  Its molecular readout
is invariant to proper rigid motions.  Handedness inversion and atom-pair
reversal invert every complete latent rotary block; with no valid stereogenic
unit, the model reduces exactly to its reflection-even graph-attention
backbone.  Finally, the $C_2$-projected ECD readout is exactly mirror even for
Number and Position and exchanges the two Symbol logits under reflection.

Each clause was proved independently above.  Together they separate the
minimal parity-sensitive input required by the no-go proposition, the
symmetry-compatible mechanism that distributes it over all atom pairs, and the
task-level projector that enforces the requested output law.

\paragraph{Scope of the guarantees.}
These proofs constrain the operator, data flow, and declared task outputs.
They do not imply that an arbitrary hidden channel has physical meaning, that
the learned field is a continuous electromagnetic field, or that algebraic
parity alone guarantees predictive accuracy.

\section{Additional Experimental Details}
\label{app:experimental-details}

\subsection{Dataset Audit}
\label{app:data-audit}

We audit splits before constructing any model cache.  Enantiomer partners must
share a partition, base identifiers must have zero overlap across partitions,
and paired parity-even tensors must agree after canonical alignment.  The
central public-ECD reconstruction additionally requires complementary R/S
labels and complementary peak signs.  Table~\ref{tab:split-audit} records the
audited split sizes used by our code.

\begin{table}[h]
  \centering
  \caption{Pair-preserving split audit.}
  \label{tab:split-audit}
  \small
  \begin{tabular}{lrrr}
    \toprule
    Dataset & Train & Validation & Test \\
    \midrule
    ChIRo R/S conformers & 326,865 & 70,099 & 69,719 \\
    ChIRo R/S base pairs & 27,542 & 5,874 & 5,840 \\
    ChIRo Ranking pairs & 24,192 & 5,184 & 5,184 \\
    public CMCDS molecules & 16,528 & 2,066 & 2,068 \\
    ACMP molecules & 952 & 120 & 120 \\
    \bottomrule
  \end{tabular}
\end{table}

For central ECD, the public archive contains fewer usable spectra than stated
in the published processed benchmark.  We do not impute missing examples or
silently mix splits.  All central-ECD tables therefore state the exact
public molecule counts they use.

\subsection{Controlled Central-ECD Retraining}
\label{app:matched-central-ecd}

To separate architecture from split construction, we retrain ECDFormer,
ChiDeK, and GSF-$\chi$ from scratch under the same preprocessing, budget, and
three-run setting.  Table~\ref{tab:central-split-matrix} reports both the
pair-preserving split used for our main comparison and the historical
sample-level split, where members of an enantiomer pair may cross partitions.
All three outputs in a row come from the same joint-spectrum checkpoint.
Table~\ref{tab:matched-central-ecd-main} summarizes the headline
pair-preserving numbers, including both the like-for-like GSF-$\chi$
default and the best prespecified configuration quoted in the main-text
Table~\ref{tab:central-results} row.

\begin{table}[t]
  \centering
  \caption{Central-ECD retraining on the same 20,662 public CMCDS
  molecules and pair-preserving split.  All three metrics come from one
  joint-spectrum checkpoint.  The final row is our best configuration.}
  \label{tab:matched-central-ecd-main}
  \small
  \setlength{\tabcolsep}{4.2pt}
  \begin{tabular}{@{}lccc@{}}
    \toprule
    Method & Position $\downarrow$ & Number $\downarrow$ & Symbol $\uparrow$ \\
    \midrule
    ECDFormer & $\meanstd{2.718}{0.065}$ & $\meanstd{1.316}{0.023}$ & $\meanstd{50.420}{0.230}$ \\
    ChiDeK & $\meanstd{2.623}{0.068}$ & $\meanstd{1.235}{0.006}$ & $\meanstd{50.264}{0.208}$ \\
    GSF-$\chi$ (ours)
      & $\bestmeanstd{2.280}{0.022}$
      & $\bestmeanstd{1.197}{0.015}$
      & $\bestmeanstd{53.102}{0.086}$ \\
    \addlinespace[1pt]
    GSF-$\chi$ (best)
      & $\meanstd{2.186}{0.015}$
      & $\meanstd{1.174}{0.004}$
      & $\meanstd{53.757}{0.315}$ \\
    \bottomrule
  \end{tabular}
\end{table}

\begin{table*}[t]
  \centering
  \caption{Central-ECD split matrix under controlled retraining.  Bold marks the
  best mean within each split and metric.}
  \label{tab:central-split-matrix}
  \scriptsize
  \setlength{\tabcolsep}{3.2pt}
  \resizebox{\textwidth}{!}{%
  \begin{tabular}{lcccccc}
    \toprule
    & \multicolumn{3}{c}{Pair-preserving 80/10/10}
    & \multicolumn{3}{c}{Sample-level 90/5/5} \\
    \cmidrule(lr){2-4}\cmidrule(lr){5-7}
    Model & Position $\downarrow$ & Number $\downarrow$ & Symbol $\uparrow$
      & Position $\downarrow$ & Number $\downarrow$ & Symbol $\uparrow$ \\
    \midrule
    ECDFormer
      & $2.718\pm0.065$ & $1.316\pm0.023$ & $50.420\pm0.230$
      & $\mathbf{1.022\pm0.027}$ & $\mathbf{0.848\pm0.029}$ & $41.669\pm1.756$ \\
    ChiDeK
      & $2.623\pm0.068$ & $1.235\pm0.006$ & $50.264\pm0.208$
      & $2.527\pm0.183$ & $1.318\pm0.131$ & $50.233\pm0.770$ \\
    GSF-$\chi$
      & $\mathbf{2.280\pm0.022}$ & $\mathbf{1.197\pm0.015}$ & $\mathbf{53.102\pm0.086}$
      & $2.058\pm0.018$ & $1.153\pm0.003$ & $\mathbf{58.314\pm0.842}$ \\
    \bottomrule
  \end{tabular}%
  }
\end{table*}

The pair-preserving split has zero train--test pair overlap and is the primary
analysis.  GSF-$\chi$ leads all three outputs there.  Under this
contract every mean gap exceeds the sum of the two reported standard
deviations, and two-sided Welch tests on the run statistics ($n{=}3$ per
method) give $p<0.01$ for Position and Symbol against both baselines and for
Number against ECDFormer; the remaining comparison, Number against ChiDeK,
gives $t{=}4.1$ ($p{=}0.034$).  The sample-level split changes the ordering
of the even metrics, illustrating why numbers from different split contracts
should not be pooled into one comparison.

\section{Extended Ablation Studies}
\label{app:ablations}

\subsection{Full Central-Chirality Mechanism Results}
\label{app:central-ablation-full}

Figure~\ref{fig:central-ablation} summarizes the main mechanism trends, with
values printed on the bars.  Each variant changes exactly one component of the
complete model; architecture, training budget, the three-run set, and the
checkpoint rule are held fixed, so the remaining differences are attributable
to the ablated component alone.  The Number, Position, and Symbol columns are
produced by one joint-ECD checkpoint shared by all central rows, not by
separately tuned task-specific models.

\subsection{Axial Ablation Setup}
\label{app:mechanism-ablation}

Table~\ref{tab:mechanism-ablation} ablates our best configuration in
Table~\ref{tab:axial-results}.  All
rows use the same ACMP split, $d=48$, $L=2$, $H_\chi/H=2/4$, optimizer, number
of epochs, three retained runs, and within-run checkpoint rule.  No ablation
variant receives its own run, checkpoint rule, or decoder.  Disabled modules
remain instantiated and have their output replaced by the stated fixed
quantity; parameter count therefore cannot explain the differences.

\begin{table*}[t]
  \centering
  \caption{Axial mechanism ablation around the exact GSF-$\chi$ headline
  configuration in Table~\ref{tab:axial-results}.  Rotation and Symbol are
  parity-sensitive; Position and Number are parity-even diagnostics.  Every
  row uses the same three runs, training budget, and evaluation procedure as
  the complete model.}
  \label{tab:mechanism-ablation}
  \small
  \setlength{\tabcolsep}{4.5pt}
  \begin{tabular}{lcccc}
    \toprule
    Variant & Rotation $\uparrow$ & Position $\downarrow$ & Number $\downarrow$ & Symbol $\uparrow$ \\
    \midrule
    GSF-$\chi$ (full) & $\mathbf{77.8\pm1.7}$ & $2.77\pm0.09$ & $1.01\pm0.03$ & $74.5\pm1.3$ \\
    Anchor queries only & $69.2\pm3.0$ & $2.80\pm0.06$ & $0.99\pm0.03$ & $67.3\pm2.6$ \\
    \addlinespace[1pt]
    w/o unit-conditioned axes & $75.3\pm1.3$ & $2.78\pm0.05$ & $0.99\pm0.02$ & $73.9\pm1.3$ \\
    + unit-conditioned frequencies & $75.6\pm1.7$ & $2.77\pm0.09$ & $1.00\pm0.02$ & $\mathbf{74.9\pm2.1}$ \\
    w/o learned pair gate & $76.9\pm4.1$ & $2.83\pm0.04$ & $1.01\pm0.05$ & $66.5\pm5.8$ \\
    w/o geometric phase initialization & $77.5\pm0.8$ & $2.81\pm0.11$ & $0.99\pm0.02$ & $74.5\pm3.8$ \\
    \midrule
    Anchor-local chirality & $71.1\pm4.9$ & $2.81\pm0.07$ & $1.01\pm0.03$ & $66.7\pm3.5$ \\
    Pooled chirality token & $71.7\pm1.7$ & $2.80\pm0.04$ & $1.01\pm0.02$ & $63.9\pm2.6$ \\
    Reflection-even backbone & $49.7\pm0.5$ & $\mathbf{2.69\pm0.07}$ & $1.00\pm0.02$ & $52.2\pm0.0$ \\
    \bottomrule
  \end{tabular}
\end{table*}

Restricting support to stereogenic-anchor queries lowers Rotation from
$77.78\%$ to $69.17\%$, a decrease of $8.61\%$.  The representation
controls are more consistent on Symbol; anchor-local attention and a pooled
chirality token lower it by $7.85\%$ and $10.61\%$, respectively.  Together
with the separately disclosed
same-checkpoint support intervention in
Table~\ref{tab:scope-intervention-results}, these results show that no single
restricted topology dominates the complete model across the parity-sensitive
outputs.  All nine rows instantiate 154,440 parameters.

The internal controls are likewise metric dependent.  Removing
unit-conditioned axes lowers Rotation by $2.50\%$ and Symbol by $0.63\%$;
adding unit-conditioned frequencies raises Symbol by $0.43\%$ but lowers
Rotation by $2.22\%$.  Removing the learned pair gate lowers Symbol by $8.07\%$, whereas removing geometric phase initialization leaves the rounded
Symbol mean unchanged.  Position and Number change much less and do not
consistently favor the full model, as expected for parity-even targets.
The reflection-even control remains at chance on Rotation; its 52.23\%
official Symbol score is exactly the 50\% real-peak baseline plus the released
evaluator's fixed correct slots for zero-peak molecules.

\begin{table*}[t]
  \centering
  \caption{Per-run values for the headline-base axial ablation.
  Rotation and ECD checkpoints use their corresponding within-run selection
  rules.  The last row uses sample standard deviation.}
  \label{tab:axial-runs}
  \small
  \resizebox{\textwidth}{!}{%
  \begin{tabular}{crrcccc}
    \toprule
    Run & Rotation epoch & ECD epoch & Rotation $\uparrow$ & Position $\downarrow$ & Number $\downarrow$ & Symbol $\uparrow$ \\
    \midrule
    A & 20 & 25 & 78.33 & 2.730 & 0.975 & 74.52 \\
    B & 24 & 25 & 75.83 & 2.703 & 1.041 & 75.80 \\
    C & 24 & 25 & 79.17 & 2.876 & 1.017 & 73.25 \\
    \midrule
    Mean $\pm$ std & -- & -- & $77.78\pm1.73$ & $2.770\pm0.093$ & $1.011\pm0.033$ & $74.52\pm1.27$ \\
    \bottomrule
  \end{tabular}
  }
\end{table*}

\begin{table*}[t]
  \centering
  \caption{Factorial separation of interaction support and ECD output
  projection on ACMP.  All configurations use the same three prespecified
  runs and checkpoint rule; the test split is evaluated only
  after restoring the selected checkpoint.  Accuracies are percentages.}
  \label{tab:topology-readout-factorial}
  \small
  \setlength{\tabcolsep}{3.5pt}
  \resizebox{\textwidth}{!}{%
  \begin{tabular}{llrcccc}
    \toprule
    Interaction & Readout & Parameters & Position $\downarrow$ & Number $\downarrow$
      & Symbol $\uparrow$ & Pair complement $\uparrow$ \\
    \midrule
    Global GSF-$\chi$ & $C_2$-projected & 154,440
      & $2.844\pm0.054$ & $1.014\pm0.021$ & $67.73\pm1.60$ & $100.00\pm0.00$ \\
    Anchor-local & $C_2$-projected & 154,440
      & $2.765\pm0.070$ & $0.994\pm0.035$ & $65.82\pm3.51$ & $100.00\pm0.00$ \\
    \addlinespace[1pt]
    Global GSF-$\chi$ & Unconstrained & 134,714
      & $2.785\pm0.036$ & $0.997\pm0.038$ & $72.51\pm0.49$ & $92.22\pm5.00$ \\
    Anchor-local & Unconstrained & 134,714
      & $2.810\pm0.018$ & $1.002\pm0.035$ & $65.07\pm2.17$ & $82.00\pm2.00$ \\
    \bottomrule
  \end{tabular}}
\end{table*}

\section{Robustness and Fine-Grained Evaluation}
\label{app:robustness}

Each subsection below pairs an experimental procedure with its results:
annotation corruption (Appendix~\ref{app:annotation-noise}), subtype and
multi-unit breakdowns (Appendix~\ref{app:subtypes}), distance-resolved field
reach (Appendix~\ref{app:field-reach}), head allocation and capacity
(Appendix~\ref{app:capacity}), and efficiency and scope
(Appendix~\ref{app:scope-of-evaluation}).  Further controls (paired
bootstrap, automatic unit detection, RotA conformer
robustness, and low-data mirror generalization) follow as separate
subsections with their own tables.

\subsection{Annotation Corruption}
\label{app:annotation-noise}

For corruption rate $r$, we draw a fixed pair-level mask from the evaluation
seed.  Half of the selected units are removed ($\rho_s=0$) and half have their
sign flipped.  The same corruption is applied to both members of a pair so that
the experiment measures annotation quality rather than accidental pair
leakage.  Rates are 2.5, 5, 7.5, and 10\%, matching the range used by ChiDeK.
We report both task performance and exact-parity residuals; the latter should
remain at numerical precision even as accuracy degrades, because the readout
law is independent of annotation correctness.

At 10\% pair-level annotation corruption, official Symbol decreases by
$0.64\%$ while Number and Position remain exactly unchanged.  The curve is
not monotone at every intermediate rate; we therefore interpret the endpoint
and per-seed values rather than fitting a trend.  All mirror residuals remain
exactly zero, confirming that an incorrect annotation can hurt accuracy
without breaking the algebraic output law.

\subsection{Axial Subtypes and Number of Units}
\label{app:subtypes}

ACMP contains chiral-atom pairs, allene-like structures, spiral atoms/chains,
biaryls, C--N, C--B, and rare C--C heterobiaryls, and nonbiaryls.  We report
sample count, real-peak count, and all four task metrics for every subtype.  A second
stratification groups molecules by one, two, or three valid stereogenic units.
This is particularly diagnostic for GSF-$\chi$ because, unlike a single-center
query, its unit contributions are defined as an order-independent sum.

Subtype and scaling rows are descriptive.  The dominant biaryl group contains
30 test pairs and reaches 81.44\% real-peak Symbol accuracy; all other subtypes
contain at most nine pairs.  In particular, the perfect multi-unit row contains only two pairs.  Molecular-size bins are likewise confounded with subtype.

\subsection{Distance-Resolved Field Reach}
\label{app:field-reach}

For every atom pair, we record the minimum graph distance of either atom from a
stereogenic anchor and aggregate $|\Delta\ell_{ij}|$ by distance.  We then mask
the correction to query-anchor, one-hop, two-hop, or global pairs without
changing parameters.  This produces a fixed-model support-sensitivity curve:
performance changes are tied directly to which atom-pair corrections remain
active.

The post-training intervention probes support without retraining; keeping
only anchor-query corrections lowers Symbol by $5.31\%$, and keeping all
anchor-incident corrections lowers it by $5.73\%$.  Aggregate Number and
Position metrics are unchanged across the three masks.  Together with the
comparable correction magnitudes at graph distances 0 through 4+, this shows
that the trained checkpoint relies on remote--remote corrections.

\begin{figure}[t]
  \centering
  \includegraphics[width=0.86\columnwidth]{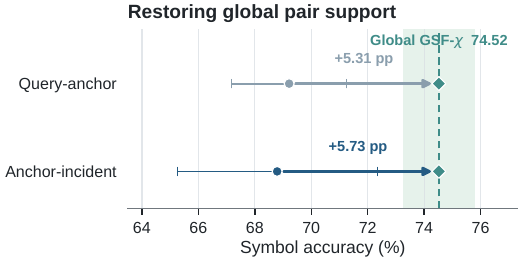}
  \caption{Same-checkpoint support intervention.  Arrows connect each
  restricted-mask mean to the all-pair GSF-$\chi$ mean.  Horizontal whiskers
  and the green band show standard deviations over three runs.}
  \label{fig:global-reach}
\end{figure}

\subsection{Head Allocation and Capacity}
\label{app:capacity}

Table~\ref{tab:capacity-full} keeps the same runs and checkpoint rule while
changing head allocation or model size.  Making every head chiral reduces Symbol by $8.28\%$, supporting a mixed even/odd attention pathway.
The 64-times larger $d=256,L=8$ model slightly improves Position but worsens
Number and loses $5.31\%$ Symbol accuracy.  The axial gain therefore cannot be
attributed to parameter count alone.

\begin{table*}[t]
  \centering
  \caption{Head allocation and capacity around our best configuration.  Every
  row uses the same runs and within-run checkpoint rule.}
  \label{tab:capacity-full}
  \small
  \begin{tabular}{lrrrr}
    \toprule
    Configuration & Parameters & Position $\downarrow$ & Number $\downarrow$ & Symbol $\uparrow$ \\
    \midrule
    $d=48,L=2$, 2/4 chiral heads
      & 154,440 & $2.770\pm0.093$ & $\mathbf{1.011\pm0.033}$ & $\mathbf{74.52\pm1.27}$ \\
    $d=48,L=2$, 4/4 chiral heads
      & 157,644 & $2.820\pm0.012$ & $1.035\pm0.017$ & $66.24\pm2.78$ \\
    $d=256,L=8$, 4/8 chiral heads
      & 9,898,102 & $\mathbf{2.734\pm0.047}$ & $1.037\pm0.047$ & $69.21\pm0.37$ \\
    \bottomrule
  \end{tabular}
\end{table*}

\begin{table*}[t]
\centering
\caption{Annotation-corruption analysis on ACMP using our best checkpoint set. Mean $\pm$ sample standard deviation over three runs.}
\label{tab:annotation-corruption-results}
\small
\begin{tabular}{ccccc}
\toprule
Corruption & Position $\downarrow$ & Number $\downarrow$ & Symbol $\uparrow$ & Real-peak Symbol $\uparrow$ \\
\midrule
0.0\% & $2.770 \pm 0.093$ & $1.011 \pm 0.033$ & $74.52 \pm 1.27$ & $73.33 \pm 1.33$ \\
2.5\% & $2.770 \pm 0.093$ & $1.011 \pm 0.033$ & $74.95 \pm 0.74$ & $73.78 \pm 0.77$ \\
5.0\% & $2.770 \pm 0.093$ & $1.011 \pm 0.033$ & $75.58 \pm 0.74$ & $74.44 \pm 0.77$ \\
7.5\% & $2.770 \pm 0.093$ & $1.011 \pm 0.033$ & $73.89 \pm 1.27$ & $72.67 \pm 1.33$ \\
10.0\% & $2.770 \pm 0.093$ & $1.011 \pm 0.033$ & $73.89 \pm 1.69$ & $72.67 \pm 1.76$ \\
\bottomrule
\end{tabular}
\end{table*}

\begin{table*}[t]
\centering
\caption{Post-training support-mask intervention over our best checkpoint set. The full model is trained globally; only the correction mask is changed at evaluation.}
\label{tab:scope-intervention-results}
\small
\resizebox{\textwidth}{!}{%
\begin{tabular}{lcccc}
\toprule
Retained correction support & Position $\downarrow$ & Number $\downarrow$ & Symbol $\uparrow$ & Real-peak Symbol $\uparrow$ \\
\midrule
anchor queries only & $2.773 \pm 0.098$ & $1.011 \pm 0.033$ & $69.21 \pm 2.05$ & $67.78 \pm 2.14$ \\
anchor-incident pairs & $2.773 \pm 0.098$ & $1.011 \pm 0.033$ & $68.79 \pm 3.55$ & $67.33 \pm 3.71$ \\
all atom pairs & $2.770 \pm 0.093$ & $1.011 \pm 0.033$ & $74.52 \pm 1.27$ & $73.33 \pm 1.33$ \\
\bottomrule
\end{tabular}
}
\end{table*}

\begin{table*}[t]
\centering
\caption{ACMP subtype analysis for our best global model over three runs. Groups with very few pairs are descriptive only.}
\label{tab:axial-subtype-results}
\small
\resizebox{\textwidth}{!}{%
\begin{tabular}{lrrrrr}
\toprule
Subtype & Pairs & Rotation $\uparrow$ & Position $\downarrow$ & Number $\downarrow$ & Real-peak Symbol $\uparrow$ \\
\midrule
Allene-like structure & 2 & $75.00 \pm 25.00$ & $4.151 \pm 0.225$ & $1.886 \pm 0.408$ & $66.67 \pm 28.87$ \\
Biaryl structure & 30 & $72.78 \pm 1.93$ & $2.215 \pm 0.176$ & $0.850 \pm 0.022$ & $81.44 \pm 2.64$ \\
Chiral atom pair & 5 & $80.00 \pm 10.00$ & $2.942 \pm 0.053$ & $0.632 \pm 0.000$ & $69.05 \pm 4.12$ \\
Heterobiaryl (C-B) & 4 & $87.50 \pm 12.50$ & $4.088 \pm 0.260$ & $0.500 \pm 0.000$ & $57.14 \pm 14.29$ \\
Heterobiaryl (C-N) & 6 & $94.44 \pm 9.62$ & $4.412 \pm 0.096$ & $1.225 \pm 0.000$ & $76.19 \pm 21.82$ \\
Nonbiaryl & 9 & $70.37 \pm 6.42$ & $2.859 \pm 0.075$ & $1.217 \pm 0.026$ & $56.94 \pm 9.62$ \\
Spiral atom and chain & 4 & $95.83 \pm 7.22$ & $2.040 \pm 0.164$ & $1.323 \pm 0.000$ & $69.23 \pm 7.69$ \\
\bottomrule
\end{tabular}
}
\end{table*}

\begin{table*}[t]
\centering
\caption{Scaling analysis for our best model on ACMP. Multi-unit results contain only two test pairs.}
\label{tab:unit-size-results}
\small
\resizebox{\textwidth}{!}{%
\begin{tabular}{llrrrrr}
\toprule
Stratification & Group & Pairs & Rotation $\uparrow$ & Position $\downarrow$ & Number $\downarrow$ & Real-peak Symbol $\uparrow$ \\
\midrule
unit count & multiple units & 2 & $100.00 \pm 0.00$ & $2.005 \pm 0.192$ & $0.000 \pm 0.000$ & $100.00 \pm 0.00$ \\
unit count & one unit & 58 & $77.01 \pm 1.80$ & $2.796 \pm 0.102$ & $1.028 \pm 0.034$ & $72.60 \pm 1.37$ \\
\midrule
atom count & 15--25 atoms & 16 & $76.04 \pm 1.80$ & $2.593 \pm 0.088$ & $0.989 \pm 0.036$ & $68.42 \pm 2.63$ \\
atom count & 26--39 atoms & 29 & $72.99 \pm 1.99$ & $3.065 \pm 0.055$ & $1.124 \pm 0.050$ & $71.23 \pm 1.37$ \\
atom count & 40+ atoms & 15 & $88.89 \pm 3.85$ & $2.388 \pm 0.224$ & $0.775 \pm 0.000$ & $82.05 \pm 4.44$ \\
\bottomrule
\end{tabular}
}
\end{table*}

\begin{table}[t]
\centering
\caption{Magnitude of the learned correction by minimum graph distance from a stereogenic anchor for one representative run of our best configuration.}
\label{tab:field-reach-results}
\small
\begin{tabular}{lrr}
\toprule
Distance & Atom-pair blocks & Mean $|\Delta\ell_{ij}|$ \\
\midrule
0 & 16048 & $0.05101$ \\
1 & 29860 & $0.04588$ \\
2 & 42248 & $0.05403$ \\
3 & 33832 & $0.05794$ \\
4+ & 47528 & $0.05597$ \\
\bottomrule
\end{tabular}
\end{table}

\begin{table}[t]
\centering
\caption{Inference efficiency on one NVIDIA H100 80GB HBM3 with batch size 32. Latency is the median of five 50-batch measurements after ten warm-up batches. All controls retain the inactive modules, so parameter counts are exactly matched. The last row is the ordinary graph transformer this work extends, obtained by disabling the signed field, and is therefore the reference for the cost of the mechanism.}
\label{tab:efficiency-results}
\small
\begin{tabular}{lrrr}
\toprule
Mechanism & Parameters & Molecules/s $\uparrow$ & Peak MiB $\downarrow$ \\
\midrule
global GSF-$\chi$ & 154,440 & $6370.3$ & $754.1$ \\
anchor-local $\chi$ & 154,440 & $14744.2$ & $172.8$ \\
pooled $\chi$ token & 154,440 & $13967.6$ & $172.8$ \\
reflection-even backbone & 154,440 & $15032.5$ & $172.8$ \\
\bottomrule
\end{tabular}
\end{table}

\subsection{Efficiency and Scope of the Evaluation}
\label{app:scope-of-evaluation}

The global field has a measurable computational cost; at batch size 32 it
processes 6,370 molecules/s and peaks at 754 MiB, compared with 15,033
molecules/s and 173 MiB for the reflection-even control, a $2.36\times$
latency and $4.36\times$ memory overhead of the direct
$|\mathcal S|N^2$ implementation, while staying below 1\,GiB at 154,440
parameters.  Every row in Table~\ref{tab:efficiency-results} uses the
$C_2$-projected ECD readout, so these measurements include the concatenated
$F(\chi)$/$F(-\chi)$ encoder batch; the comparison therefore isolates field
topology while measuring end-to-end projected inference.  On achiral
molecules the operator reduces exactly to that backbone
(Eq.~\ref{eq:guarantee-achiral}), so the cost is paid only where stereogenic
units exist.  Beyond cost, the empirical scope has two boundaries: central
ECD uses the reproducible public subset rather than ChiDeK's unreleased
processed set, and the implementation assumes a supplied stereogenic unit
set, though handedness within a unit is computed from the conformer
(Appendix~\ref{app:canonicalization-audit}) and
Section~\ref{sec:results-generalization} measures the cost of removing that
remaining input.  These boundaries affect empirical scope and uncertainty,
not the exact parity identities.

\subsection{End-to-End Cost of the $C_2$ Projection}
\label{app:c2-efficiency}
The projected implementation concatenates $\chi$ and $-\chi$ into one
accelerator batch. We benchmark the complete ECD forward, including Number,
Position, and Symbol heads, against the one-sign unconstrained model.
Timing uses 20 warm-up iterations followed by seven synchronized repetitions.

\begin{table*}[t]
\centering
\caption{End-to-end inference cost on one NVIDIA H100 80GB. Mean $\pm$ sample
standard deviation over three checkpoints.}
\label{tab:c2-projection-efficiency}
\small
\resizebox{\textwidth}{!}{%
\begin{tabular}{lrrrrr}
\toprule
Readout & Parameters & Batch-1 latency (ms) & Batch-1 peak MiB & Batch-32 molecules/s & Batch-32 peak MiB \\
\midrule
one-sign unconstrained & 134,714 & $3.432\pm0.015$ & $65.7\pm0.0$ & $8264.4\pm18.1$ & $391.6\pm0.0$ \\
$C_2$-projected & 154,440 & $4.056\pm0.006$ & $66.3\pm0.0$ & $6370.3\pm6.6$ & $754.1\pm0.0$ \\
\bottomrule
\end{tabular}}
\end{table*}

At batch size one, projection changes latency by $1.18\times$; at batch size
32, its peak allocation is $1.93\times$ the one-sign readout and throughput
decreases from $8264$ to $6370$ molecules/s. These measurements include the
concatenated two-sign encoder evaluation and therefore quantify the deployment
cost omitted by an encoder-only parameter count.

\subsection{Pair-Level Uncertainty for Parity-Odd Targets}
\label{app:pair-bootstrap}

The dispersion reported elsewhere in this paper is taken across training runs
and describes optimization variability at a fixed test set.  We additionally
quantify uncertainty with respect to sampling of the test set itself, by
jointly resampling the same 60 complete ACMP enantiomer pairs for GSF-$\chi$
and ChiDeK over $10{,}000$ bootstrap replicates.  Resampling whole
enantiomer pairs rather than individual molecules keeps each pair's two
configurations together, and the complete reported run sets are aggregated
before each paired difference is computed.  Unlike a comparison between
marginal confidence intervals, this procedure removes the pair-specific
difficulty shared by the two methods, which is why we report the paired
difference itself rather than two separate intervals.

\begin{table}[t]
  \centering
  \caption{Paired bootstrap on the two parity-odd ACMP targets against
  retrained ChiDeK.  Differences are percentage points; intervals are $95\%$
  pair-bootstrap percentile intervals obtained by resampling the same complete
  enantiomer pairs for both methods.  Positive values favor GSF-$\chi$.}
  \label{tab:parity-odd-paired-bootstrap}
  \small
  \begin{tabular}{lcc}
    \toprule
    Target & Paired $\Delta$ & $95\%$ CI \\
    \midrule
    Rotation & $\mathbf{+9.67}$ & $[+1.83,\,+17.83]$ \\
    Symbol   & $\mathbf{+7.96}$ & $[+0.60,\,+15.24]$ \\
    \bottomrule
  \end{tabular}
\end{table}

Both intervals exclude zero.  For Rotation, a two-sided paired bootstrap test
gives $p=0.0154$ and a pair-cluster sign-flip randomization test gives
$p=0.0251$.  This comparison therefore statistically resolves the
advantage of GSF-$\chi$ on both parity-odd outputs, molecular Rotation and ECD
Symbol, which are exactly the two targets whose value reverses under
handedness inversion.  The parity-even targets are unaffected, as expected.
This complements the exact architectural parity guarantees of
Section~\ref{sec:method-guarantees} with finite-test-set empirical evidence.

\subsection{Automatic Stereogenic-Unit Detection}
\label{app:auto-unit-detection}

Handedness within a unit is already computed from the conformer rather than
read from an annotation (Appendix~\ref{app:canonicalization-audit}), so the
remaining external input is the stereogenic unit set itself.  We therefore
evaluate whether GSF-$\chi$ can operate with units extracted automatically
from molecular structure.  We apply the ChiralFinder/RDKit extraction
pipeline~\citep{shi2026unifying,rdkit}, an external tool developed
independently of this work, and match each detected unit against the curated
ACMP unit by its complete atom-index set.  Optical-rotation and ECD labels are
never used during extraction.

Only the atom set of each unit is taken from the extractor.  Handedness is
recomputed from the conformer under our own canonical gauge exactly as in the
curated setting, so the extractor's chirality descriptor enters neither the
model input nor the evaluation.  This keeps the substitution a test of unit
localization alone, and prevents the detector from supplying the very
pseudoscalar whose effect is being measured.

\begin{table*}[t]
  \centering
  \caption{Automatic stereogenic-unit detection on all 1,192 ACMP molecular
  configurations.  Precision, recall, and F1 are micro-averaged over exact
  atom-index-set matches against the 1,280 curated units.}
  \label{tab:auto-unit-detection}
  \small
  \begin{tabular}{lc}
    \toprule
    Detection statistic & Value \\
    \midrule
    True positive units  & 1,186 \\
    False positive units & 1,600 \\
    False negative units & 94 \\
    Precision            & 42.57\% \\
    Recall               & 92.66\% \\
    F1 score             & 58.34\% \\
    Exact molecule-level unit set & 50.00\% \\
    No detected unit     & 4.70\% \\
    Mean detected units per molecule & 2.34 \\
    \bottomrule
  \end{tabular}
\end{table*}

The extractor therefore supplies a deliberately noisy, high-recall annotation
setting, recovering $92.66\%$ of curated units while proposing roughly twice
as many candidates in total.  We then evaluate the curated-trained checkpoints
directly under these automatically extracted units, with no retraining, no
adaptation, and no change to the operator.

Under this substitution Rotation is $68.61\pm3.76\%$ and Symbol is
$62.00\pm3.51\%$, drops of $5.56$ and $9.34$ points from the same checkpoints
evaluated with curated units; the
parity-even targets are essentially unchanged at $2.823\pm0.065$ Position and
$1.016\pm0.049$ Number, as expected for quantities the signed field does not
orient.  Two observations follow.  First, both parity-odd outputs stay far
above the $50\%$ floor, so the operator degrades gracefully rather than
collapsing when its unit set is wrong roughly half the time at the molecule
level.  Second, the Rotation accuracy obtained with fully automatic extraction
still exceeds the $65.17\%$ that ChiDeK attains with curated
annotations.

We do not claim an annotation-free pipeline.  Detection precision of $42.57\%$
is the limiting factor, and axial-unit perception is an open problem in its
own right, orthogonal to the operator studied here.  What these results
establish is that the parity-sensitive signal GSF-$\chi$ extracts is a
property of the geometry and the interaction topology rather than of curated
labels, and that improvements in upstream detection translate directly into
end-to-end gains without any change to the model.

\subsection{Conformer Robustness on RotA}
\label{app:rota-conformer-robustness}

\paragraph{Evaluation setting.}
RotA contains 3,140 conformers of 650 axially chiral
molecules~\citep{shi2026unifying}.  Because ACMP is derived from this
collection, RotA is \emph{not} an independent test corpus; we use it instead to
test whether predictions remain stable across alternative conformations of the
same held-out molecular identities.  We freeze the ACMP-trained
checkpoints of each method (five per baseline, three for GSF-$\chi$) and
evaluate every RotA conformer associated with
the ACMP test set, without retraining, checkpoint selection, or parameter
fitting on RotA.  Each conformer is paired with a physically reflected copy,
while its Rotation and ECD targets are inherited from the corresponding ACMP
enantiomer pair.  This gives 283 complete physical mirror pairs (566 samples)
drawn from 59 held-out molecules, which remain the independent sampling unit.
One additional test molecule is excluded because its singleton
stereogenic-unit annotation does not define the two-endpoint geometric gauge
needed for a physical axial reflection audit.

Table~\ref{tab:rota-main} in the main text reports the accuracies.  GSF-$\chi$
degrades gracefully under the conformational shift, reaching $73.1\%$ Rotation
and $72.8\%$ Symbol on RotA conformers, against $77.8\%$ and $74.5\%$ on ACMP
(Table~\ref{tab:axial-results}).  Its margin over the matched ChiDeK
retraining meanwhile widens, from $+12.6$ and $+7.9$ points on ACMP to
$+15.4$ and $+10.7$ points here, and Number RMSE
improves by $0.07$.  GSF-$\chi$ also holds the lowest Position RMSE
($2.78$ vs.\ $2.80$), a parity-even target where no signed advantage is
expected.  That the advantage grows
rather than shrinks on unseen conformations is what one expects if the signal
is carried by geometry rather than by conformer-specific memorization.

ECDFormer produces identical outputs for a molecule and its mirror image, which
forces its Rotation accuracy to exactly chance and pins its Symbol score at the
evaluator floor with zero variance across runs; the reflection diagnostics
below confirm this directly.

Table~\ref{tab:rota-main} in the main text isolates the parity
projection rather than the encoder.  Both GSF-$\chi$ rows come from one shared representation and
one checkpoint; the two heads differ only in whether the $C_2$ projection of
Section~\ref{sec:method-readout} is applied.  The projected Symbol head
complements on every one of the 283 mirror pairs, with zero variance across
all three GSF-$\chi$ runs, whereas the unprojected Rotation head flips on
$78.8\%$ of
them, statistically indistinguishable from ChiDeK's $78.7\%$.  A head trained
without the projection therefore behaves like a conventional classifier even
when it reads a parity-correct representation, which is direct evidence that
exact mirror consistency comes from the projection and cannot be obtained by
learning alone.  Rotation flip should accordingly be read as a diagnostic, not
as a guarantee the architecture makes.

This law is realized through the complete coordinate-reflection
pipeline on every evaluated conformer, not only under an internal sign-flip
intervention; across all 283 pairs, the parity-even GSF inputs match exactly and
every supplied pseudoscalar changes sign after reflection.

\paragraph{Implementation audit.}
The released ChiralFinder annotations are recovered for $88.05\%$ of the
curated axes in this conformer set.  Native ChiralFinder preprocessing
completes for 556 of 566 samples; for the remaining ten samples ($1.77\%$) we
use the same deterministic coordinate-based fallback as in the ACMP evaluation
adapter (Appendix~\ref{app:auto-unit-detection}).  All detected, nondegenerate
mirror pairs reverse their determinant sign.

\subsection{Training-Time Support-Budget Controls}
\label{app:equal-support-retraining}

We retrain every support rather than masking a dense checkpoint.  Random and
graph-distance-matched supports retain exactly the query-anchor entry count
for each molecule and stereogenic unit.  The masks are label independent,
parity even, and shared by the two members of every enantiomer pair.  The
hard-top-$k$ variants select the same parity-even pair gate at three support
budgets; they are included as exploratory negative controls.

\begin{table*}[t]
  \centering
  \caption{Training-time support controls on ACMP.  Accuracies are percentages;
  every row uses the same three runs and checkpoint rule.}
  \label{tab:equal-support-retraining}
  \small
  \setlength{\tabcolsep}{3.8pt}
  \begin{tabular}{lrrrrr}
    \toprule
    Support & Density & Rotation $\uparrow$ & Position $\downarrow$ & Number $\downarrow$ & Symbol $\uparrow$ \\
    \midrule
    Query-anchor & 6.5\% & $56.11\pm1.73$ & $2.793\pm0.043$ & $1.019\pm0.031$ & $70.06\pm2.30$ \\
    Random-global & 6.5\% & $67.22\pm1.92$ & $2.841\pm0.013$ & $1.027\pm0.012$ & $69.64\pm6.99$ \\
    Distance-matched global & 6.5\% & $66.39\pm0.96$ & $2.817\pm0.010$ & $1.011\pm0.010$ & $65.82\pm3.21$ \\
    \addlinespace[1pt]
    Hard top-$k$ global & 6.5\% & $50.00\pm0.00$ & $2.785\pm0.060$ & $1.026\pm0.063$ & $52.87\pm0.64$ \\
    Hard top-$k$ global & 15.0\% & $51.11\pm0.48$ & $2.780\pm0.052$ & $1.013\pm0.062$ & $53.08\pm0.74$ \\
    Hard top-$k$ global & 30.0\% & $51.39\pm1.27$ & $2.814\pm0.013$ & $0.991\pm0.044$ & $52.87\pm0.64$ \\
    Full global & 100.0\% & $75.83\pm2.20$ & $2.817\pm0.052$ & $1.024\pm0.038$ & $68.37\pm1.95$ \\
    \bottomrule
  \end{tabular}
\end{table*}

At the identical 6.5\% budget, moving support from anchor queries to
random-global and distance-matched global pairs improves Rotation by $11.11\%$ and
$10.28\%$, respectively.  Full support adds another $8.61$--$9.44\%$.
Thus the Rotation gain is not explained by retaining more entries alone; the
ability to place signed interactions away from the anchor matters.  Symbol
does not share this ordering (query-anchor is comparable to random-global and
higher than distance-matched global), so this control supports a
target-specific global-placement claim rather than universal dominance on
every output.  The hard top-$k$ selector collapses near chance at all three
budgets.  We treat this as an optimization failure of discrete support
selection, not evidence against dense global interaction, and do not use it
for the mechanism claim.

\subsection{Low-Data Mirror Generalization}
\label{app:low-data-mirror}

The pair-fraction setting retains complete pair identities.  In contrast,
one-side supervision labels one configuration per training and validation
pair and withholds its mirror from both optimization and checkpoint selection.
The member is selected by a target-blind hash of pair identity.  The original
ACMP test pairs remain pair-disjoint from training in every setting.

\begin{table*}[t]
  \centering
  \caption{Pair-disjoint ACMP test performance under reduced ECD supervision.
  Symbol and complement are percentages.}
  \label{tab:low-data-mirror}
  \small
  \setlength{\tabcolsep}{3.8pt}
  \begin{tabular}{llrrrr}
    \toprule
    Labels & Readout & Position $\downarrow$ & Number $\downarrow$ & Symbol $\uparrow$ & Complement $\uparrow$ \\
    \midrule
    25\% pairs & projected & $2.830\pm0.111$ & $1.065\pm0.000$ & $67.30\pm2.65$ & $100.00\pm0.00$ \\
    25\% pairs & unconstrained & $2.799\pm0.104$ & $1.065\pm0.000$ & $58.81\pm2.67$ & $41.78\pm20.79$ \\
    50\% pairs & projected & $2.826\pm0.116$ & $1.065\pm0.000$ & $69.00\pm2.05$ & $100.00\pm0.00$ \\
    50\% pairs & unconstrained & $2.790\pm0.046$ & $1.065\pm0.000$ & $64.23\pm2.43$ & $73.11\pm4.73$ \\
    one side & projected & $2.840\pm0.105$ & $1.065\pm0.000$ & $69.00\pm2.24$ & $100.00\pm0.00$ \\
    one side & unconstrained & $2.835\pm0.033$ & $1.054\pm0.015$ & $67.62\pm4.78$ & $70.89\pm7.34$ \\
    100\% pairs & projected & $2.844\pm0.054$ & $1.014\pm0.021$ & $67.73\pm1.60$ & $100.00\pm0.00$ \\
    100\% pairs & unconstrained & $2.785\pm0.036$ & $0.997\pm0.038$ & $72.51\pm0.49$ & $92.22\pm5.00$ \\
    \bottomrule
  \end{tabular}
\end{table*}

\begin{table}[t]
  \centering
  \caption{Mirror completion after one-enantiomer supervision.  Withheld
  training mirrors are never used for optimization or checkpoint selection.
  Symbol and complement are percentages.}
  \label{tab:one-side-mirror-completion}
  \small
  \begin{tabular}{lrr}
    \toprule
    Readout & Mirror Symbol $\uparrow$ & Complement $\uparrow$ \\
    \midrule
    projected & $77.65\pm1.70$ & $100.00\pm0.00$ \\
    unconstrained & $64.45\pm2.58$ & $77.34\pm7.52$ \\
    \bottomrule
  \end{tabular}
\end{table}

Projection is most useful when mirror supervision is missing; it improves
withheld-mirror Symbol accuracy by $13.20\%$ and enforces exact pair
complement.  On pair-disjoint test molecules, its gain is largest at 25\%
paired supervision ($8.49\%$), decreases at 50\% ($4.78\%$), and
reverses with complete paired supervision, where the unconstrained head has
enough labels to learn the relation approximately.  This is the expected
inductive-bias trade-off rather than a universal raw-accuracy gain.

\section{Comparison and Analysis Procedures}
\label{app:selection-regimes}

Table~\ref{tab:central-results} reports our best
benchmark settings so that each task has one complete headline comparison;
the ECD columns of ECDFormer and ChiDeK in Table~\ref{tab:central-results}
are matched retrainings, described in Section~\ref{sec:results-main}.  The
matched comparisons in Appendices~\ref{app:matched-central-ecd}
and~\ref{app:matched-axial-results} use prespecified run sets and
hyperparameters, retraining the baselines on identical molecules, splits,
budgets, and evaluators.

\subsection{Controlled Axial Retraining}
\label{app:matched-axial-results}

We additionally retrain ECDFormer and ChiDeK on the same 1,192
ACMP molecules, pair-preserving split, evaluator, and 25-epoch budget.
Table~\ref{tab:matched-axial-results} reports the retrained baselines
alongside our best GSF-$\chi$ configuration; per-run values are serialized in
the released machine-readable manifests.  Two-sided Welch tests give
$p{=}4.9\times10^{-4}$ for Rotation, $p{=}5.8\times10^{-4}$ for Symbol, and
$p{=}0.014$ for Number RMSE; Position is tied at the reported precision.

\begin{table}[t]
  \centering
  \caption{ACMP retraining with a common split, budget, and checkpoint
  rule; baselines average five runs.  The final row is our best
  configuration.  Rotation and Symbol are percentages.}
  \label{tab:matched-axial-results}
  \scriptsize
  \setlength{\tabcolsep}{2.0pt}
  \resizebox{0.5\columnwidth}{!}{%
  \begin{tabular}{@{}lcccc@{}}
    \toprule
    Method & Rotation $\uparrow$ & Position $\downarrow$ & Number $\downarrow$ & Symbol $\uparrow$ \\
    \midrule
    ECDFormer & $\meanstd{50.00}{0.59}$ & $\meanstd{3.696}{0.244}$ & $\meanstd{1.139}{0.086}$ & $\meanstd{52.77}{0.14}$ \\
    ChiDeK & $\meanstd{65.17}{1.60}$ & $\bestmeanstd{2.770}{0.097}$ & $\meanstd{1.118}{0.055}$ & $\meanstd{66.62}{1.41}$ \\
    \addlinespace[1pt]
    GSF-$\chi$ (best) & $\bestmeanstd{77.78}{1.73}$ & $\bestmeanstd{2.770}{0.093}$ & $\bestmeanstd{1.011}{0.033}$ & $\bestmeanstd{74.52}{1.27}$ \\
    \bottomrule
  \end{tabular}}
\end{table}

\subsection{Continuous Axial-Rotation Construction}
\label{app:rotation-construction}

The analysis in Figure~\ref{fig:axial-rotation-analysis} uses ACMP pairs 27
and 601 because they already serve as the biaryl and non-biaryl qualitative
examples; trajectory separation was not a selection criterion.  For each
molecule, we remove the canonically oriented axial bond, choose the connected
component on its second endpoint, and rotate that component about the bond by
$0^\circ,20^\circ,\ldots,340^\circ$.  Each of the 18 conformers then passes
through the ordinary unit extraction, role canonicalization, feature builder,
and one locked axial-ECD encoder.  Both series contain nine conformers of each
configuration.  We compute $z^+$ and $z^-$ before dimensionality reduction;
UMAP uses cosine distance, five neighbors, minimum distance $0.08$, and a fixed
random state.  The polar curves are not projected; they are raw cosine
similarities to the $0^\circ$ representation.  Their machine-readable manifest
contains every angle, extracted sign, embedding, and similarity.

\section{Case-Study Selection and Additional Examples}
\label{app:case-selection}

\begin{figure*}[t]
  \centering
  \includegraphics[width=0.96\textwidth]{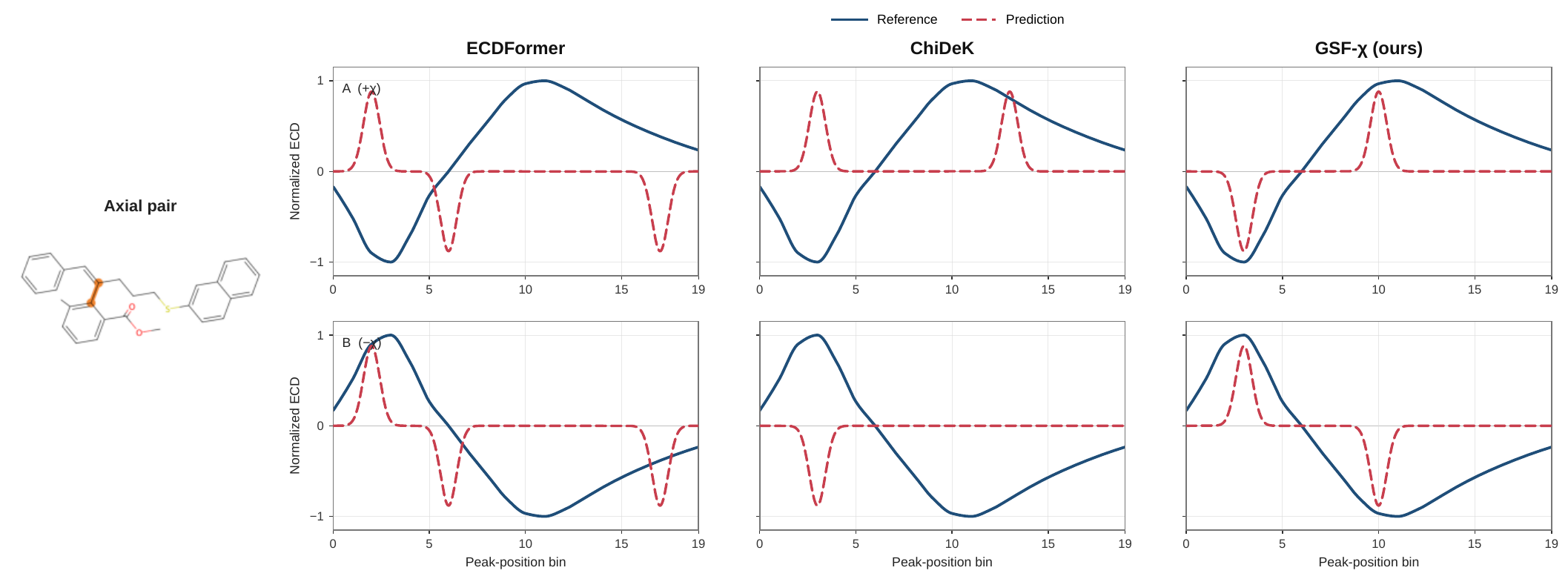}
  \caption{Axial-ECD prediction for an enantiomer pair.  Rows are opposite
  configurations; solid and dashed curves are reference and prediction.
  GSF-$\chi$ recovers Number and both Symbols, with a one-bin displacement of
  the second peak; the baselines make visibly different count or sign errors.}
  \label{fig:ecd-prediction-example}
\end{figure*}

Figure~\ref{fig:ecd-prediction-example} makes the paired output law visible on
ACMP pair 601.  Both configurations have $N=2$ and $P=(3,11)$, with Symbols
$(-,+)$ and $(+,-)$.  ECDFormer repeats the same three-peak prediction across
the pair; ChiDeK predicts two peaks with $(+,+)$ in the first row but only one
peak in the second; GSF-$\chi$ preserves the paired law, predicts both Symbols
correctly, and places the peaks at $(3,10)$, one bin from the reference.

The pair is chosen by a deterministic post-hoc rule.  We first retain pairs for
which the locked GSF-$\chi$ checkpoint predicts Number and Symbol exactly in
both configurations, obeys the complete pair law, and misses no Position by
more than one bin.  We then maximize ChiDeK's combined Number and Symbol
errors, breaking ties by total component error and base ID.  This rule affects
only the illustration, never model selection or aggregate metrics.  This
appendix also records the attribution procedure and adds multi-unit and failure
cases.

\begin{figure*}[t]
  \centering
  \includegraphics[width=\textwidth]{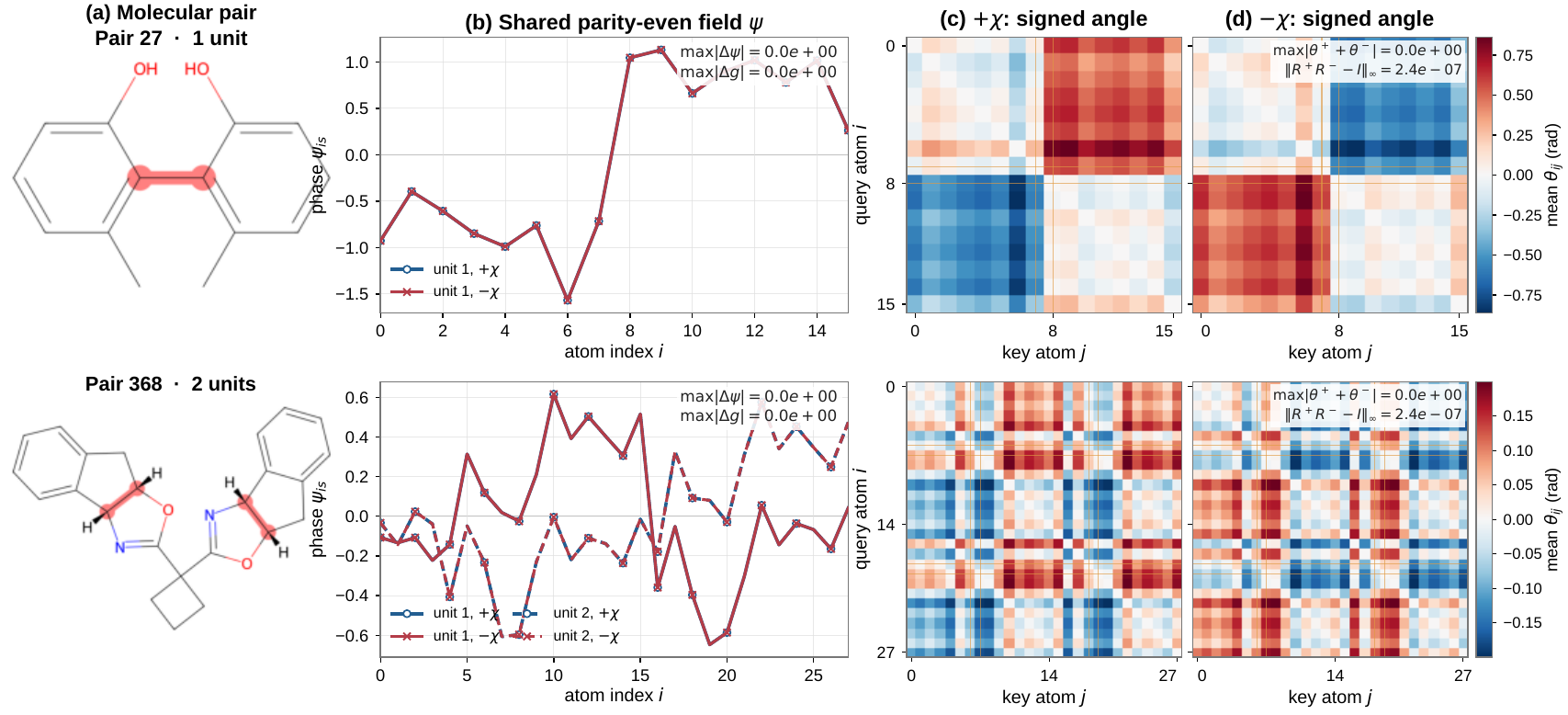}
  \caption{Configuration inversion inside GSF-$\chi$.  Rows show one- and
  two-unit molecules.  \textbf{(a)} Molecular structure; \textbf{(b)} shared
  parity-even field; \textbf{(c,d)} signed-angle maps for opposite
  configurations.  The residuals verify operator inversion to numerical
  precision; atom-pair indices are not physical rotation angles.}
  \label{fig:parity-field-comparison}
\end{figure*}

Figure~\ref{fig:parity-field-comparison} opens the operator and shows that
inversion preserves the parity-even field and gates while negating every
signed latent angle, including for a two-unit molecule.  This operator-level
view is the visual counterpart of Section~\ref{sec:method-guarantees}.

\begin{figure*}[t]
  \centering
  \includegraphics[width=\textwidth]{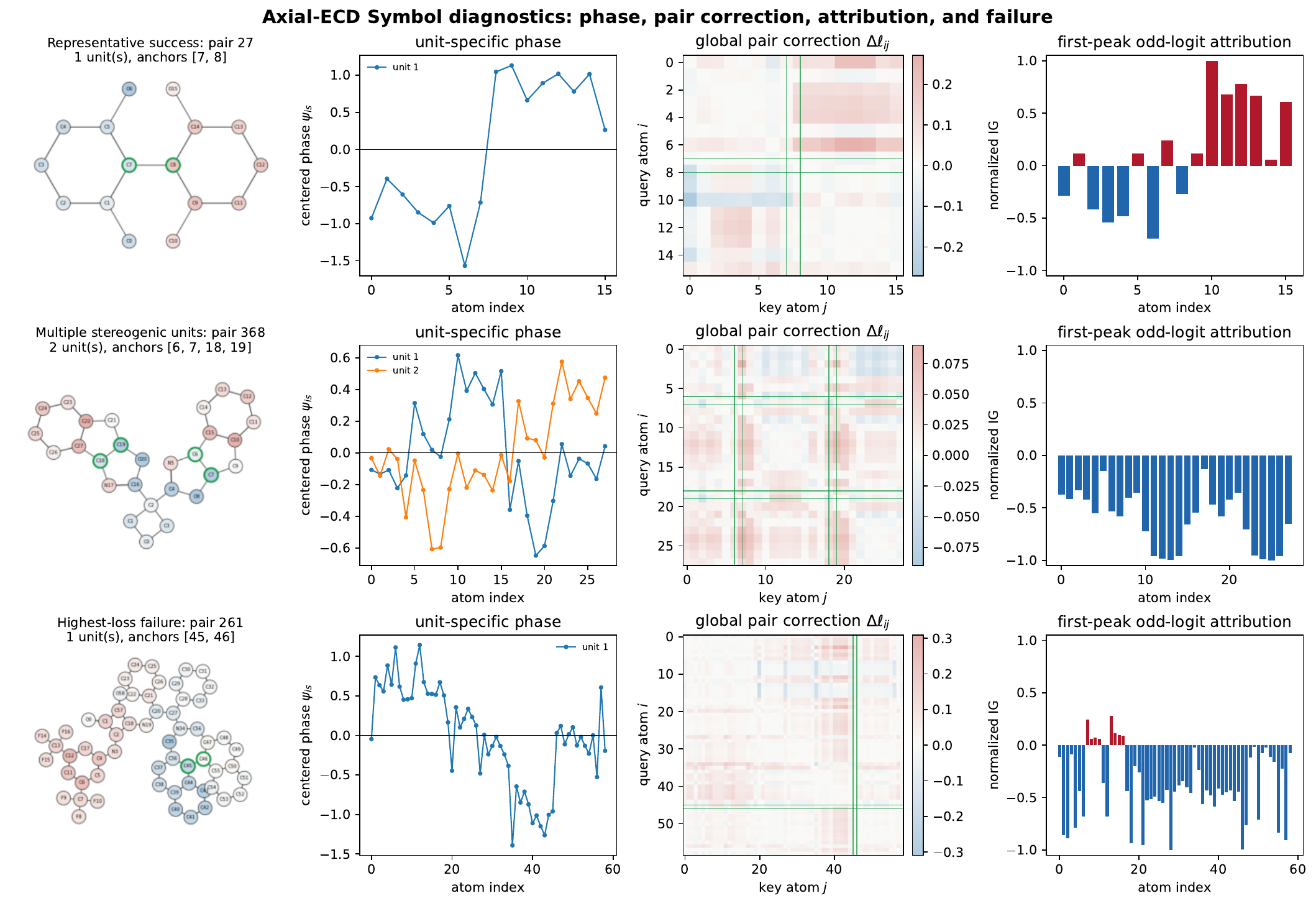}
  \caption{Detailed axial-ECD Symbol diagnostics.  Rows contain the
  deterministic success (pair 27), multi-unit example (pair 368), and
  highest-loss failure (pair 261).  Columns show the phase-colored structure,
  unit-specific phase curves, all-atom rotary correction, and integrated
  gradients for the first Symbol logit.  These panels are qualitative
  diagnostics; the same-checkpoint mask intervention provides a separate
  support-sensitivity test.}
  \label{fig:case-study-diagnostics}
\end{figure*}

Every molecular panel is generated from serialized model outputs and an RDKit
molecule; structures are not manually redrawn.  Atom colors encode the
centered phase, and green outlines identify stereogenic anchors.  The compact
case figure prioritizes direct paired Symbol predictions.  The detailed diagnostic
view in Figure~\ref{fig:case-study-diagnostics} retains unit-specific phase
curves and integrated-gradient attributions using the zero odd representation
as baseline and 64 interpolation steps.  It also contains the deterministically
selected highest-loss failure, rather than hiding it in a success-only figure.
The companion table summarizes even predictions and paired Symbol correctness,
while the machine-readable manifest stores every
$(\text{position},\text{symbol})$ tuple.  Table~\ref{tab:axial-subtype-results}
complements these examples with exhaustive subtype-level statistics.

\begin{figure*}[t]
  \centering
  \includegraphics[width=\textwidth]{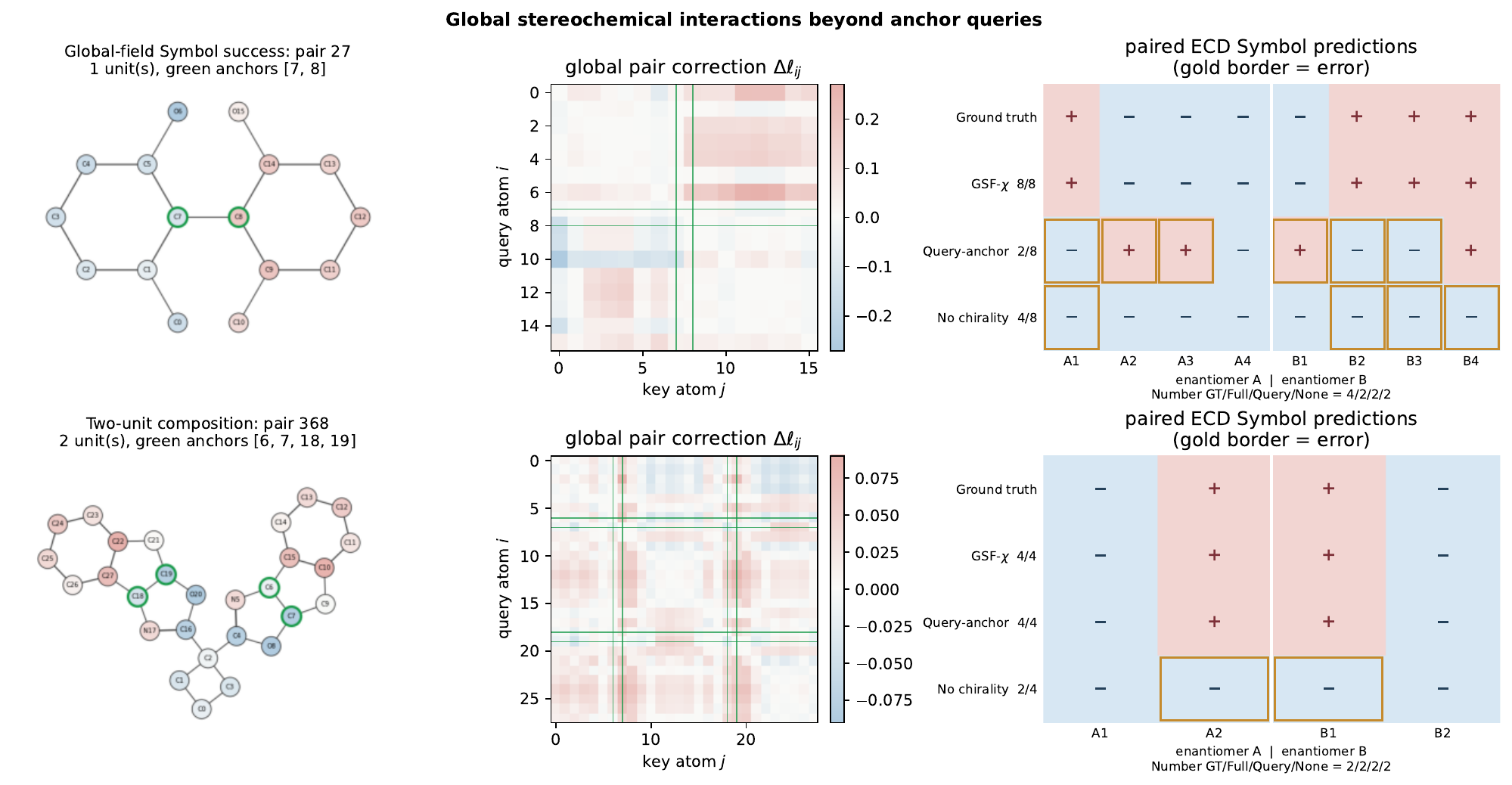}
  \caption{Axial-ECD Symbol cases.  Rows show a control-separating Symbol
  success and a two-unit example.  Columns show the phase-colored molecule,
  all-atom rotary correction, and paired Symbol predictions; gold borders
  mark errors.}
  \label{fig:case-study}
\end{figure*}

Pair 261 is the largest paired Symbol-NLL failure under the same locked
checkpoint.  GSF-$\chi$, query-anchor, and no-chirality classify 2/6, 2/6,
and 3/6 real-peak Symbols, respectively.  We report the failure without
assigning an unsupported chemical explanation that the experiment cannot identify;
its broad field and attribution demonstrate that global reach alone does not
guarantee correctness.

\begin{table*}[t]
  \centering
  \caption{Predictions for the deterministic cases.  Number and Position are
  exactly shared by each reflected pair; Symbol reports correctly classified
  real peaks across both members.  $P_{\mathrm{cmp}}$ contains the first
  $\min(N,\widehat N)$ positions used by the released evaluator; the manifest
  contains every individual signed tuple.}
  \label{tab:case-predictions}
  \scriptsize
  \setlength{\tabcolsep}{3pt}
  \resizebox{\textwidth}{!}{%
  \begin{tabular}{llccc}
    \toprule
    Pair / role & Ground truth & GSF-$\chi$ & Query-anchor & No chirality \\
    \midrule
    27 / Symbol success
      & $N=4,\ P=(3,7,11,16)$
      & $N=2,\ P_{\mathrm{cmp}}=(3,11),\ 8/8$
      & $N=2,\ P_{\mathrm{cmp}}=(3,10),\ 2/8$
      & $N=2,\ P_{\mathrm{cmp}}=(5,11),\ 4/8$ \\
    368 / two units
      & $N=2,\ P=(3,9)$
      & $N=2,\ P_{\mathrm{cmp}}=(3,11),\ 4/4$
      & $N=2,\ P_{\mathrm{cmp}}=(3,11),\ 4/4$
      & $N=2,\ P_{\mathrm{cmp}}=(5,11),\ 2/4$ \\
    261 / highest loss
      & $N=3,\ P=(1,4,12)$
      & $N=3,\ P_{\mathrm{cmp}}=(3,10,13),\ 2/6$
      & $N=3,\ P_{\mathrm{cmp}}=(3,9,13),\ 2/6$
      & $N=3,\ P_{\mathrm{cmp}}=(3,9,13),\ 3/6$ \\
    \bottomrule
  \end{tabular}
  }
\end{table*}

\subsection{Additional Paired ECD Predictions}
\label{app:additional-ecd-cases}

\begin{figure*}[p]
  \centering
  \includegraphics[width=\textwidth]{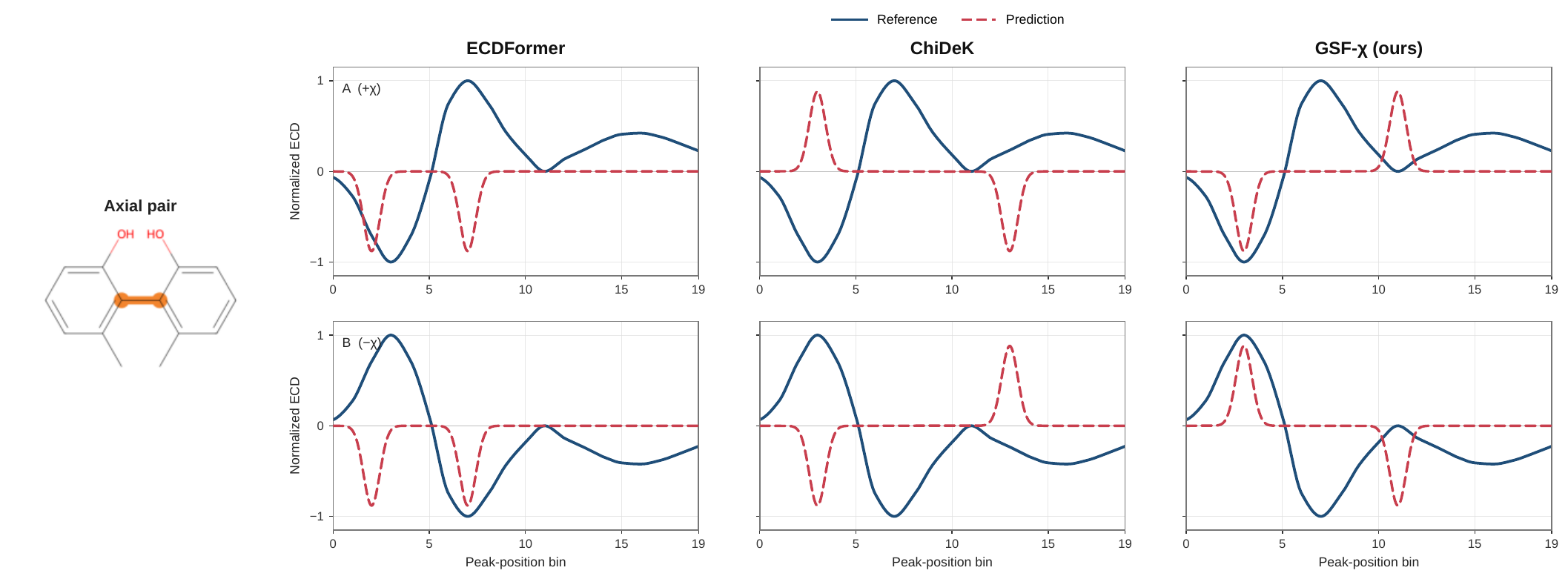}
  \caption{Control-separating Symbol case (ACMP pair 27).  GSF-$\chi$ predicts
  complementary signs at Positions $(3,11)$ and matches the corresponding
  released signs in both configurations.  It predicts two rather than four
  peaks, so this is deliberately a Symbol-focused case rather than exact
  spectrum recovery; Table~\ref{tab:case-predictions} evaluates all four
  real-peak slots and contrasts the scope controls.}
  \label{fig:ecd-case-success}
\end{figure*}

\begin{figure*}[p]
  \centering
  \includegraphics[width=\textwidth]{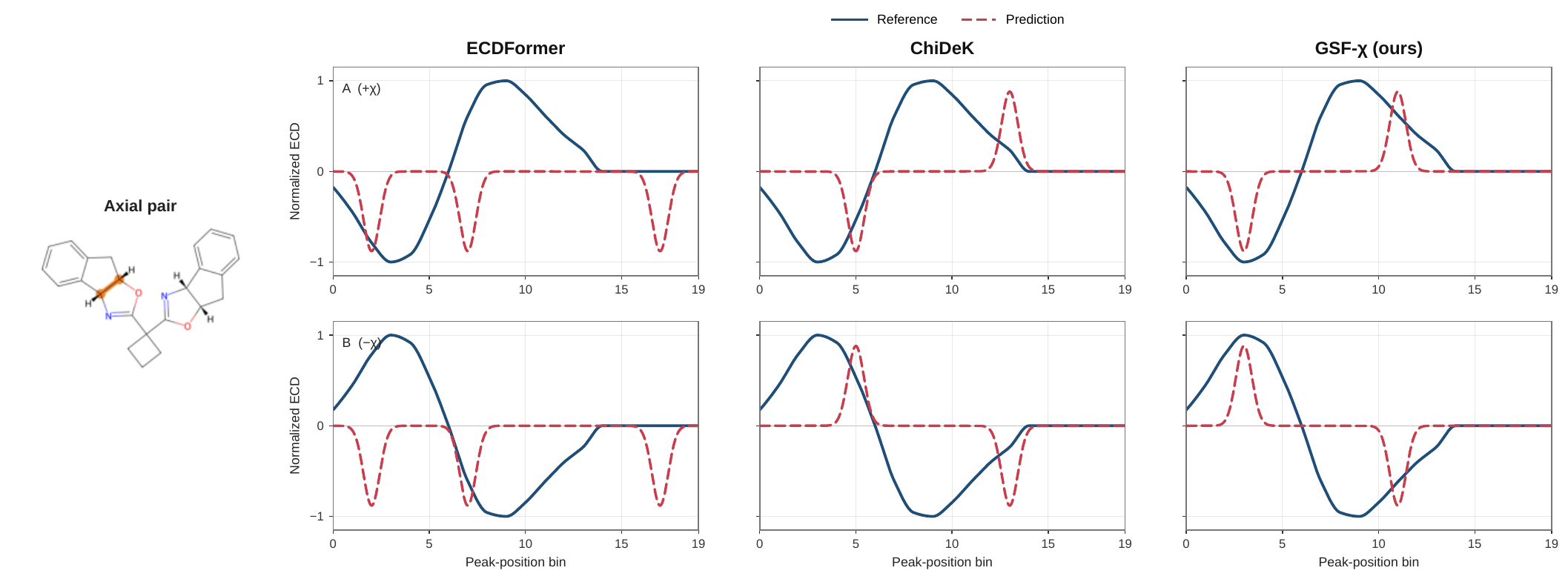}
  \caption{Two-unit case (ACMP pair 368).  GSF-$\chi$ preserves Number and
  Position across the reflected pair and predicts both complementary Symbols;
  its second Position is shifted from 9 to 11.  The example demonstrates that
  two stereogenic fields can contribute without imposing an order over units,
  not that every spectral component is exact.}
  \label{fig:ecd-case-368}
\end{figure*}

\begin{figure*}[p]
  \centering
  \includegraphics[width=\textwidth]{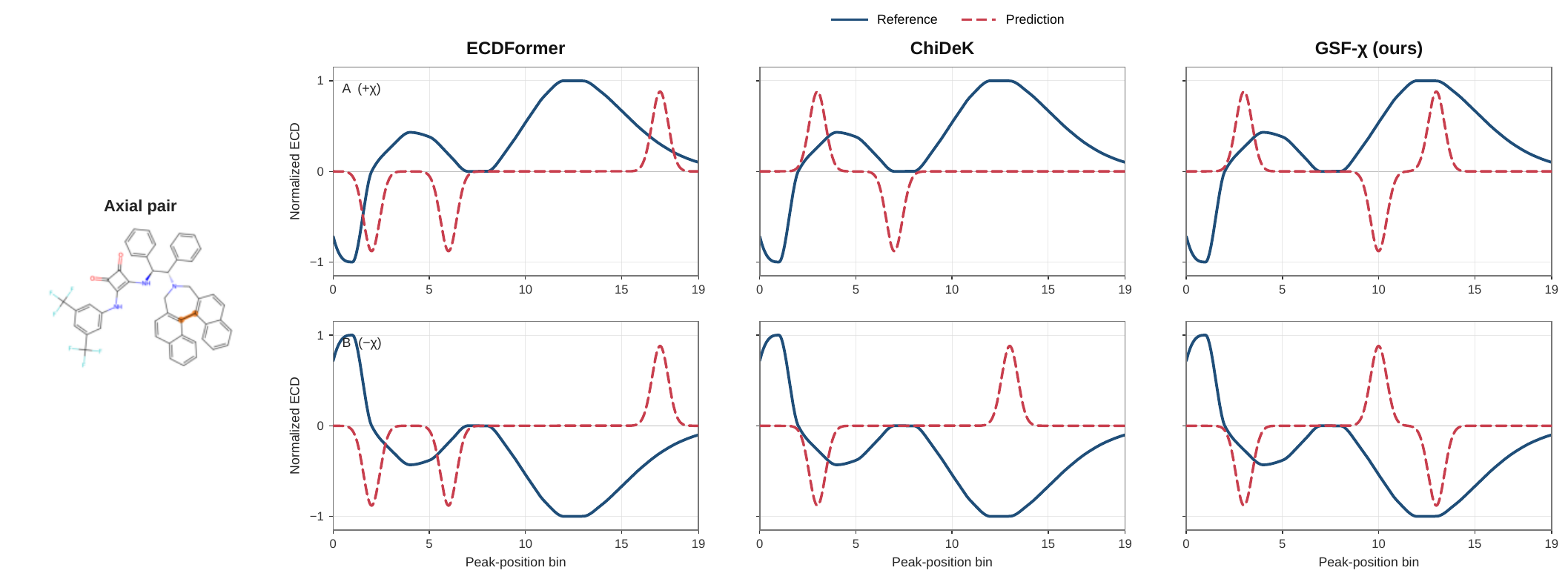}
  \caption{Locked highest-loss failure (ACMP pair 261).  GSF-$\chi$ obeys its
  exact mirror law, with Number and Positions agreeing across configurations and
  the predicted Symbols complementing, but the prediction is inaccurate for both
  Positions and two of three Symbols.  Algebraic consistency constrains the
  error; it does not guarantee chemical accuracy.}
  \label{fig:ecd-case-261}
\end{figure*}

Figures~\ref{fig:ecd-case-success}--\ref{fig:ecd-case-261} use the same plotting
contract as the main paired example; each model column receives the identical
reference sequence within a row, and dashed curves render only predicted peak
number, position, and symbol at fixed amplitude.  Pairs 27, 368, and 261 are
the deterministic Symbol success, multi-unit example, and highest-loss failure
under one representative visualization checkpoint.

\end{document}